\documentclass{article}

\PassOptionsToPackage{numbers, compress}{natbib}

\usepackage[preprint]{neurips_2026}

\usepackage[utf8]{inputenc} % allow utf-8 input
\usepackage[T1]{fontenc}    % use 8-bit T1 fonts
\usepackage{hyperref}       % hyperlinks
\hypersetup{colorlinks=true,linkcolor=blue,citecolor=blue,urlcolor=blue}
\usepackage{url}            % simple URL typesetting
\usepackage{booktabs}       % professional-quality tables
\usepackage{amsfonts}       % blackboard math symbols
\usepackage{nicefrac}       % compact symbols for 1/2, etc.
\usepackage{microtype}      % microtypography
\usepackage{xcolor}         % colors
\usepackage{multirow}       % multi-row table cells
\usepackage{graphicx}       % figures
\usepackage{float}          % [H] fixed float placement
\usepackage{eso-pic}        % absolute page positioning

\title{FilmBench: A Film-Grade Benchmark for Cinematic Video Generation}

\author{%
  \normalfont
  Shengyi Wang\textsuperscript{$\dag$1,2}\enspace
  Niantong Li\textsuperscript{$\dag$1}\enspace
  Guangzheng Hu\textsuperscript{$\dag$1}\enspace
  Hong Qi\textsuperscript{*3}\enspace
  Fei Ding\textsuperscript{1,2}\enspace
  Weixu Qiao\textsuperscript{1}\\[2pt]
  Jinlin Wang\textsuperscript{1}\enspace
  Xiaotong Lv\textsuperscript{1}\enspace
  Peng Han\textsuperscript{1}\enspace
  Zimeng Li\textsuperscript{1}\enspace
  Fanshu Ding\textsuperscript{1}\enspace
  Yushu Wang\textsuperscript{1}\\[2pt]
  Han Wu\textsuperscript{1}\enspace
  Jingjing Chen\textsuperscript{1}\enspace
  Chongxiao Wang\textsuperscript{1,2}\enspace
  Yanhao Wu\textsuperscript{1,2}\enspace
  Chenglong Huang\textsuperscript{1,2}\\[2pt]
  Xiaoqian Zhu\textsuperscript{1,2}\enspace
  Jie Tian\textsuperscript{1,2}\enspace
  Hua Li\textsuperscript{1,2}\enspace
  Jingjing Fan\textsuperscript{1,2}\enspace
  Mingshuang Tang\textsuperscript{3}\\[2pt]
  Zhong Li\textsuperscript{3}\enspace
  Hengxia Qiang\textsuperscript{3}\enspace
  Weibin Chen\textsuperscript{1}\enspace
  Jinyang Zhen\textsuperscript{1}\\[2pt]
  Bing Zhao\textsuperscript{1}\enspace
  Lin Qu\textsuperscript{1}\enspace
  Jing Li\textsuperscript{*1,2}\enspace
  Hu Wei\textsuperscript{*1}
  \AND
  \normalfont
  \textsuperscript{1}Alibaba Group\\[1pt]
  \textsuperscript{2}Moku Lab, Hujing Digital Media \& Entertainment Group\\[1pt]
  \textsuperscript{3}Beijing Film Academy\\[6pt]
  \parbox{0.92\linewidth}{\centering
  \textsuperscript{$\dag$}~Equal contribution.\quad \textsuperscript{*}~Corresponding authors.\\[2pt]
  \texttt{qihong@bfa.edu.cn},\; \texttt{kongwang@alibaba-inc.com},\; \texttt{lj225205@alibaba-inc.com}\\[6pt]
  \raisebox{-0.5\height}{\includegraphics[height=0.2in]{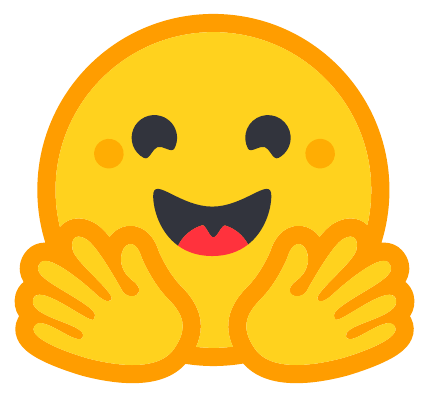}}\;\raisebox{-4pt}{\href{https://huggingface.co/datasets/skylenage/FilmBench}{\textcolor{blue}{\texttt{https://huggingface.co/datasets/skylenage/FilmBench}}}}\\
  \raisebox{-0.5\height}{\includegraphics[height=0.2in]{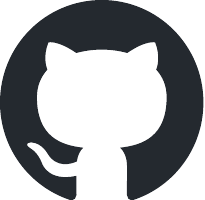}}\;\raisebox{-2pt}{\href{https://github.com/Neo-yk/FilmOps}{\textcolor{blue}{\texttt{https://github.com/Neo-yk/FilmOps}}}}
  }
}

\begin{document}
\raggedbottom

\maketitle
% footnotes removed — shown inline in author block

% --- Logos: single centered row (ali, dt, youku, hujing) ---
\AddToShipoutPicture*{%
  \put(0,\LenToUnit{10.1in}){%
    \makebox[\paperwidth]{%
      \raisebox{-0.5\height}{\includegraphics[height=0.28in]{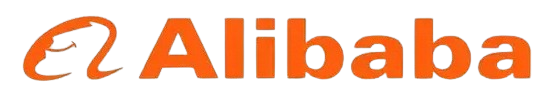}}\hspace{3.2in}%

      \raisebox{-0.5\height}{\includegraphics[height=0.22in]{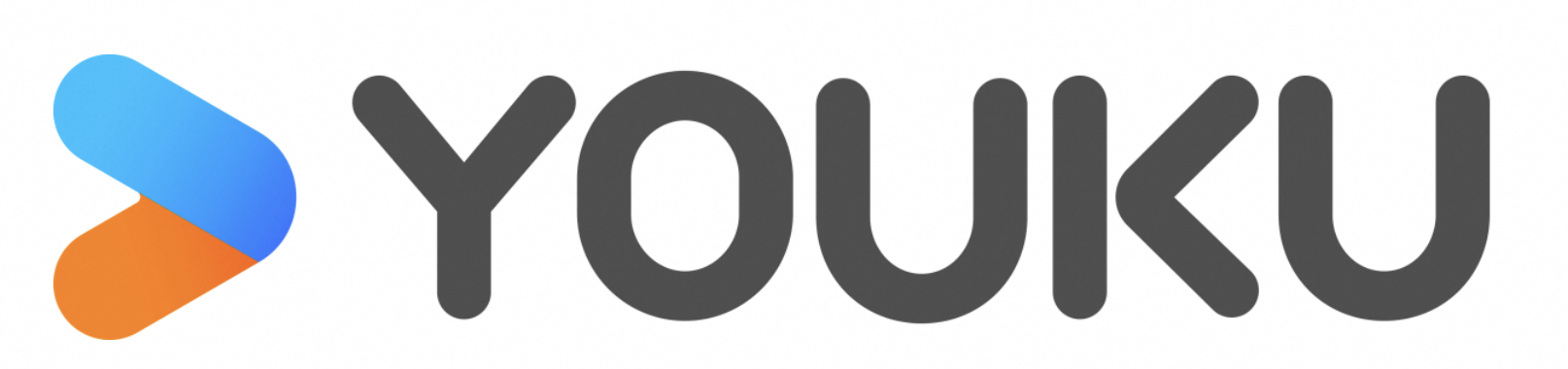}}%
    }%
  }%
}

% ----- Abstract -----
% !TeX root = neurips_2026.tex
% ===========================================================
% Abstract of FilmBench
% ===========================================================
\begin{abstract}
Progress in video generation keeps narrowing the visual gap between
AI-generated and professionally produced footage, yet most benchmarks still
draw prompts from web sources or LLM templates and score them with untrained,
generic multimodal models. More fundamentally, their evaluation taxonomies
remain rudimentary (overall visual quality, coarse text alignment and temporal
smoothness) rather than the professional Cinematic Language criteria by which
films are actually made and judged, so they assess basic video plausibility
rather than film-grade craft. We introduce \textbf{FilmBench},
a text-to-video (T2V) and reference-to-video (R2V) benchmark grounded in
the professional Cinematic Language of the film-academy tradition and
co-developed with directors and faculty from the Beijing Film Academy and the
Hujing Digital Media \& Entertainment Group film studio. It rests on three choices. First, prompts are \emph{reverse-engineered} from clips of
award-winning films spanning 20 cinematic genres and chosen by professional
directors, so every prompt is
anchored to a verified live-action reference; the prompts follow real shot
lists, and most script multiple shots (1{,}056 of the 1{,}169 prompts are
multi-shot), unlike prior single-clip benchmarks. Second, evaluation follows a
three-level Cinematic taxonomy of \textbf{3 axes, 12 components and 35
(T2V) +3 (R2V-only) sub-metrics}. Third, we develop an in-house expert-grade automatic
evaluation agent and open-source its core suite of Cinematic Language operators
(FilmOps). Benchmarking leading video generation models (9 for T2V, 7 for R2V), the evaluator
reproduces the human model ranking at model-level Spearman $\rho = 0.95$ (T2V)
and $0.96$ (R2V). Scores fall well below prior web-style benchmarks, with two
consistent gaps in \emph{dynamic aesthetics} and a marked \emph{single- to
multi-shot} performance drop that widens for weaker models.
\end{abstract}

% ----- Main Sections of FilmBench -----
% !TeX root = neurips_2026.tex
% ===========================================================
% Section: Introduction
% ===========================================================
\section{Introduction}
\label{sec:intro}

\begin{figure}[!htbp]
\centering
\includegraphics[width=\linewidth]{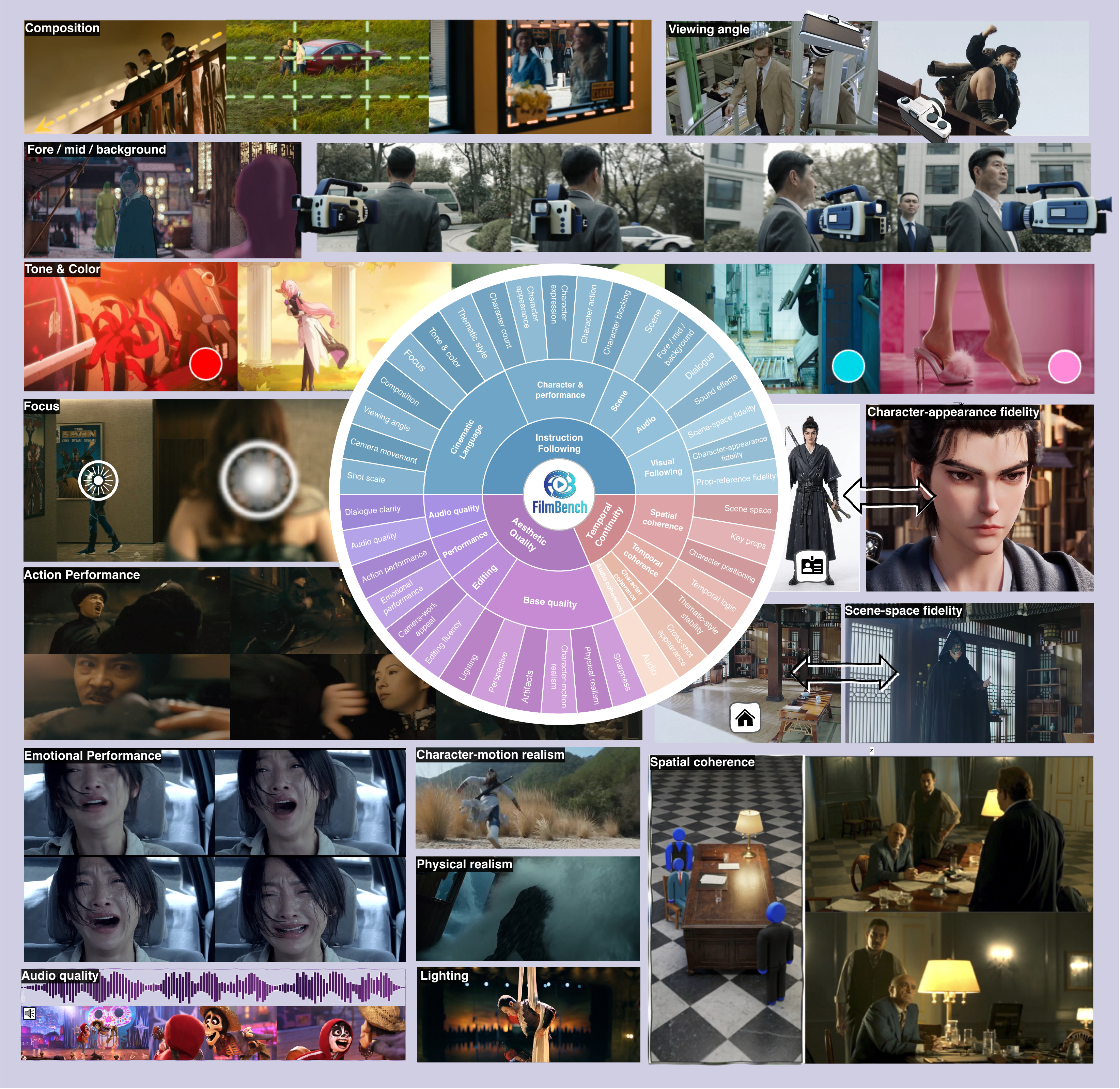}
\caption{FilmBench evaluation taxonomy: 3~L1 axes, 12~L2 components, 35{+}3 (R2V-only) L3 sub-metrics, built from clips across 20 film genres curated by expert directors.}
\label{fig:dim_framework}
\end{figure}

\begin{figure}[!htbp]
\centering
\includegraphics[width=\linewidth]{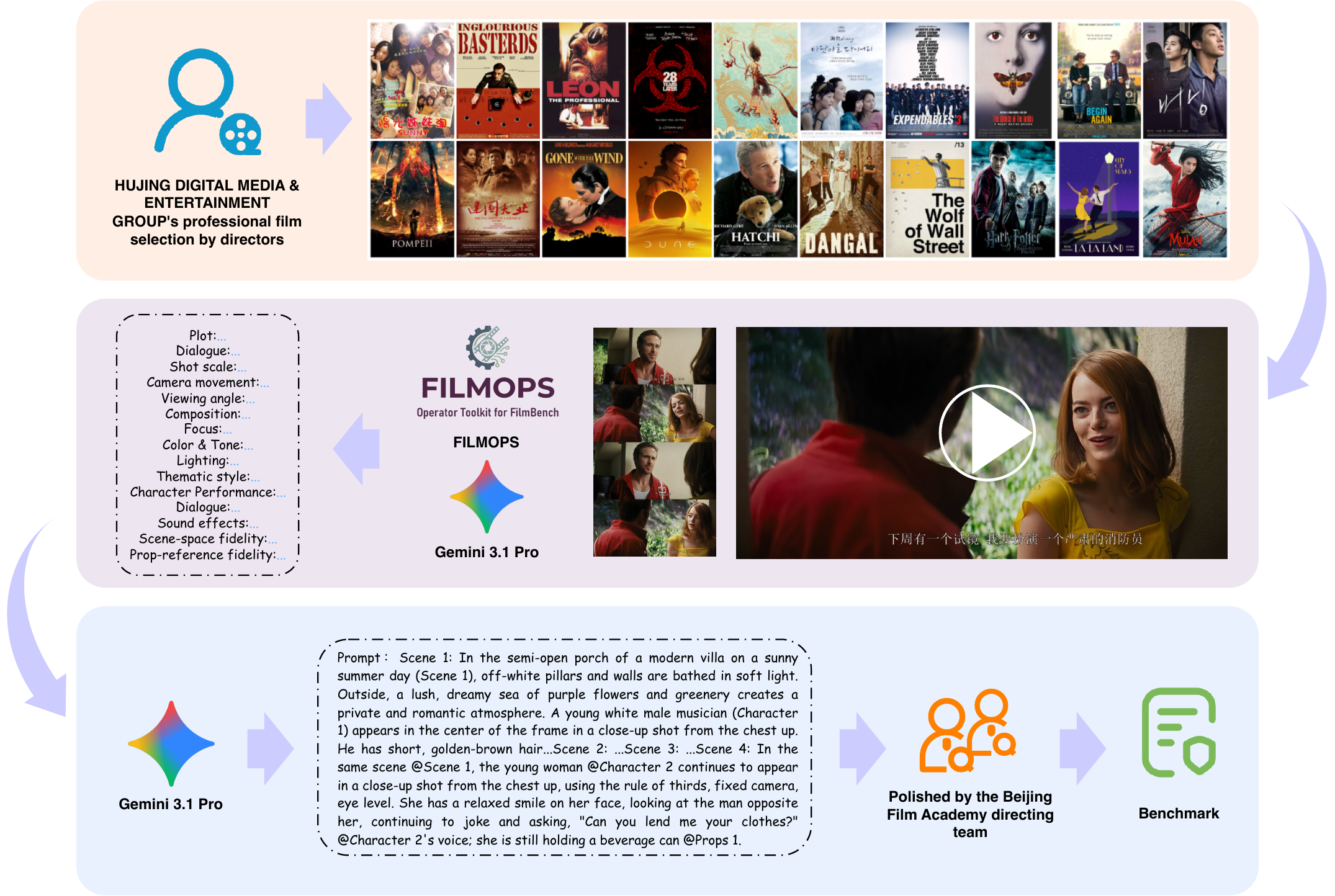}
\caption{The director-driven reverse-engineering pipeline that turns
award-winning film clips into film-grade prompts; see
\S\ref{sec:dim:pipeline} for details.}
\label{fig:construct_pipe}
\end{figure}

Video generation has progressed at an unprecedented pace. Diffusion- and
DiT-based models (Veo~\citep{google2025veo3,wiedemer2025video}, Kling~\citep{klingteam2025klingomni}, Seedance~\citep{seedance2026seedance20}, Hailuo~\citep{minimax2025hailuo23}, HappyHorse~\citep{happyhorse2026v10,happyhorse2026v11} and others) now
produce $\sim$15-second photorealistic clips with controllable
camera and even synchronized audio, blurring the line between AI-generated
footage and professionally produced cinema. A natural next
question is whether the capability of these models has reached the
pass mark of \emph{professional cinematic creation}. Modern film production
follows a rigorous Cinematic Language (shot scale, camera
movement, shot perspective, lighting, scene blocking, composition and
film-grade audio design) refined over more than a century of practice.
Existing benchmarks were not designed with this professional,
academy-taught Cinematic Language in mind, and so far they paint only a
coarse, optimistic picture of model capability.

\paragraph{The current benchmarking landscape, and what is missing.}
The community has produced a rich body of evaluation suites in the past
three years. Foundational multi-dimensional benchmarks such as
VBench~\citep{huang2024vbench}, VBench-2.0~\citep{huang2025vbench2} and
Video-Bench~\citep{han2025videobench} dissect quality into 9--18 generic
dimensions, and learned scorers such as
VideoScore~\citep{he2024videoscore} push automatic evaluation closer to
human preference. Compositional and fine-grained benchmarks
(T2V-CompBench~\citep{sun2024t2vcompbench}, FETV~\citep{liu2023fetv}) target
attribute, motion and interaction binding; temporal, motion and multi-shot
benchmarks (ChronoMagic-Bench~\citep{yuan2024chronomagic},
VMBench~\citep{ling2025vmbench}, SLVMEval~\citep{slvmeval2026},
MSVBench~\citep{msvbench2026}) probe long-horizon, motion and cross-shot
fidelity;
reasoning, physical and social benchmarks
(TiViBench~\citep{tivibench2026}, SVBench~\citep{svbench2026},
WorldJen~\citep{worldjen2026}, RBench~\citep{rbench2026}) stress higher-order
or embodied capabilities; the I2V line
(ConsistI2V/I2V-Bench~\citep{ren2024consisti2v},
UI2V-Bench~\citep{ui2vbench2025}, IP-Bench~\citep{ipbench2026}) focuses on
image-conditioned generation; and audio--video and aesthetic benchmarks
(AVGen-Bench~\citep{avgenbench2026}, VGA-Bench~\citep{vgabench2026}) address
audio quality and visual aesthetics, respectively.

\begin{figure}[h]
\centering
\includegraphics[width=\linewidth]{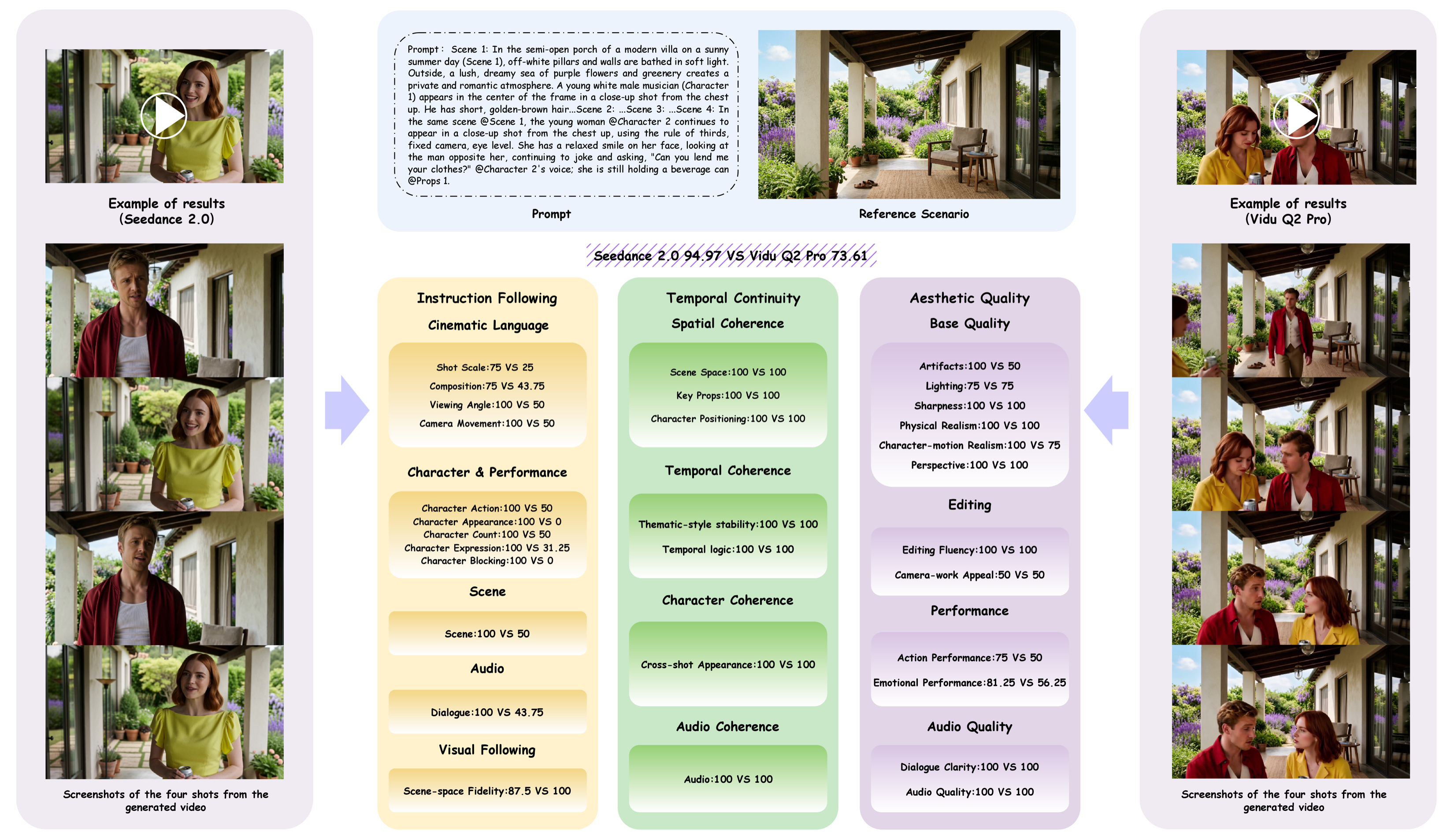}
\caption{Fine-grained, per-dimension FilmBench evaluation on a multi-shot
reference-scene (R2V) example reverse-engineered from \emph{La~La~Land}:
Seedance~2.0 (94.97) vs.\ Vidu~Q2~Pro (73.61). Discussed in the text.}
\label{fig:eval_example}
\end{figure}

Despite this breadth, none of these benchmarks evaluates models against the
standards used in real cinematic creation. They share three structural blind
spots. First, prompts are sourced from web users, captioners or LLM
templates rather than from verified professional shots, so the input
distribution drifts away from cinematic content. Second, the implicit content
distribution skews toward generic, English, web-style scenes, with no
balanced coverage of cinematic genres or culturally diverse film traditions.
Third, the evaluation taxonomies leave out cinematic axes such as shot
scale, camera movement, lighting, visual style, character performance and
audio quality, or collapse them into a single ``aesthetic'' score. As a
consequence, current leaderboards saturate while professional cinematic
creators readily reject the same outputs as not film-grade.

\paragraph{FilmBench.}
We address this gap with FilmBench, an evaluation benchmark for cinematic
text-to-video (T2V) and reference-to-video (R2V) generation, built jointly
with directors and faculty from the Beijing Film Academy and the Hujing
Digital Media~\& Entertainment Group film studio. FilmBench is grounded in
three design principles.

Figure~\ref{fig:dim_framework} gives an overview of the FilmBench evaluation
taxonomy. Expert directors from the Beijing Film Academy selected clips
spanning 20 film categories---covering the breadth of professional cinematic
genres---and used them as the foundation for both the prompt set and the
fine-grained evaluation hierarchy. The taxonomy radiates from the three L1
axes through progressively finer L2 components and L3 sub-metrics, while the
surrounding film examples illustrate the kinds of cinematic-language challenges
each dimension targets. We next describe how these expert-curated clips are
turned into structured, film-grade prompts.

(P1)~\emph{Reverse-engineered from real films} (Figure~\ref{fig:construct_pipe}): instead of authoring prompts
in the abstract, professional directors select clips from award-winning films
that stress Cinematic Language and visual expression, reverse-infer draft prompts
with a strong multimodal model (Gemini~3.1~Pro), and refine them into structured
shot-level prompts that explicitly encode scene, role, prop, shot scale,
camera movement, composition, lighting, dialogue and performance, so that
every prompt is anchored to a verified professional reference and every
fine-grained criterion in the taxonomy is covered. Because these prompts
follow real shot lists, most are multi-shot (1{,}056 of the 1{,}169 prompts),
unlike the single-clip prompts of prior benchmarks.
(P2)~\emph{Academy-aligned cinematic taxonomy}: our evaluation
dimensions follow the Cinematic Language teaching system of the Beijing
Film Academy, organized as \emph{three} first-level axes (L1):
\emph{Instruction Following}, \emph{Temporal Continuity} and
\emph{Aesthetic Quality}. These decompose into 12 second-level components (L2) and
35 third-level sub-metrics (L3). T2V and R2V share this main framework; the
R2V task additionally adds a \emph{Visual Following} L2 component of three L3
sub-metrics (scene-space, character-appearance and prop-reference
fidelity), each evaluated against the conditioning reference image, giving
R2V 13 L2 and 35+3 L3 (per prompt, only the sub-metric matching its reference
type is scored).
(P3)~\emph{Expert-grade automatic evaluator}: every sub-metric is scored by an
in-house, professional film-grade evaluation agent whose core Cinematic Language
operator suite (FilmOps) we open-source; its model-level ranking reproduces
expert film-industry rankings at Spearman $\rho{=}0.95$ (T2V) / $0.96$ (R2V)
(\S\ref{sec:dim}, \S\ref{sec:eval}).

\paragraph{A worked example.}
Figure~\ref{fig:eval_example} shows FilmBench scoring a multi-shot
reference-scene (R2V) example, a four-shot clip reverse-engineered from the
musical \emph{La~La~Land}, on Seedance~2.0 (94.97) and Vidu~Q2~Pro (73.61). Because
every L3 sub-metric is scored independently, the evaluation exposes behavior
that an aggregate score hides. Although Seedance~2.0 leads overall by a wide
margin, its \emph{scene-space fidelity} is slightly lower (87.5 vs.\ 100):
Vidu~Q2~Pro over-adheres to the reference background image and scores
lower on nearly every other dimension. It drops markedly on Cinematic Language
following (shot scale 75
vs.\ 25, viewing angle 100 vs.\ 50, camera movement 100 vs.\ 50, composition
75 vs.\ 43.75) and, having sacrificed the shot/reverse-shot staging to lock
onto the scene, exhibits character-positioning drift that drives its
\emph{character blocking} to 0. FilmBench can thus objectively credit a single
dimension (reference fidelity) while still penalizing the broader loss of
professional Cinematic Language; Appendix~\ref{app:examples} discusses further
per-dimension case studies.

\paragraph{Findings.}
We benchmark leading video-generation models (9 for T2V, 7 for R2V),
including Seedance~2.0, HappyHorse~1.1/1.0, Kling~3.0, Kling~3.0~Omni, Veo~3.1,
Grok~Imagine~Video~\citep{xai2026grok}, Vidu~\citep{bao2024vidu}, and Hailuo~2.3, and distill five field-wide
findings. \textbf{(i)~No saturation:} no model approaches the ceiling on
either task (top scores of 88.93 for T2V and 86.66 for R2V), in sharp
contrast with the near-ceiling numbers on prior web-style benchmarks; and
crucially, what separates the leaders is not generic quality but the
professional Cinematic Language sub-metrics of Instruction Following (camera
movement, shot scale, viewing angle, focus and composition) together with the
Aesthetic Quality axis, which is exactly where FilmBench resolves differences
that web-style benchmarks miss.
\textbf{(ii)~A field-wide dynamic-aesthetics bottleneck:} across all models
the lowest-scoring sub-metrics are action performance, camera-work appeal,
character-motion realism and emotional performance, whereas static
image-quality sub-metrics (sharpness, lighting) score markedly higher:
today's models render clean frames but still struggle with believable motion and
performance. \textbf{(iii)~Multi-shot is the hardest regime:} on T2V, moving
from single- to multi-shot prompts costs $7.9$ points on average and up to
$22.8$ for the weakest model, concentrated in the Cinematic Language and editing
demands of Instruction Following (framing, camera work and cut fluency); it is
multi-shot staging, rather than single-take quality, where current models fall
short. \textbf{(iv)~Reference conditioning stresses rather than reorders the
field:} across scene, prop and character references the R2V ranking stays
stable, yet conditioning lowers scores to varying degrees and penalizes the
weakest model most, and the visual-following leader is not necessarily the
overall leader, so reference behavior is best read through ranking shifts
rather than absolute per-type scores.
\textbf{(v)~No model wins on all L3 sub-metrics:} Seedance~2.0, the overall
leader, claims only 18/35 (T2V) or 20/38 (R2V) L3 championships, concentrated
on Cinematic Language sub-metrics (Cinematic Language, editing appeal and performance),
while HappyHorse~1.1 dominates 11/35 (T2V) or 10/38 (R2V) sub-metrics on
character, audio and scene dimensions; notably the three R2V-specific
visual-following championships all go to the HappyHorse family; 4~(T2V) and 2~(R2V) models do not claim any championship. This ``championship mismatch'' reveals structural
complementarity concealed by a single leaderboard number.

\paragraph{Contributions.}
We make four contributions. (i) We release \textbf{FilmBench}, a
cinematic-grade T2V/R2V benchmark whose prompts are reverse-engineered from
award-winning real films and curated jointly with directors and faculty from
the Beijing Film Academy and a professional film studio, ensuring coverage of
every fine-grained cinematic criterion. (ii) We propose an academy-aligned
evaluation taxonomy that translates professional Cinematic Language into a
three-level hierarchy of 3 L1 axes, 12 L2 components and 35 L3 sub-metrics
(35+3 with the R2V-specific Visual-Following extension), scored by an expert-grade
automatic evaluator. (iii) We open-source \textbf{FilmOps}, a modular
Cinematic Language operator suite with trained weights and inference scripts,
enabling the community to build customizable, expert-knowledge-driven judge
agents for cinematic video evaluation. (iv)
We conduct a comprehensive cinematic evaluation over leading T2V/R2V models,
show the automatic evaluator reproduces expert rankings at $\rho{=}0.95$~(T2V)
/ $0.96$~(R2V), and expose dynamic aesthetics as the shared bottleneck, discussing
directions for future research.

% !TeX root = neurips_2026.tex
% ===========================================================
% Section: Related Work
% ===========================================================
\section{Related Work}
\label{sec:related}

We trace how video-generation evaluation has evolved through three stages (from
generic quality scoring, to capability-specific stress tests, to increasingly
film-like assessment) and show that each stage, in closing one gap, exposes
the next, until the accumulated gaps motivate FilmBench precisely.

\paragraph{From distribution metrics to multi-dimensional, human-aligned evaluation.}
Early evaluation repurposed action-recognition datasets (UCF-101, MSR-VTT,
Kinetics) and reported single scalars such as FVD/IS, which neither localize
\emph{which} aspect of a video fails nor align well with human perception. As
diffusion- and DiT-based generators---including the systems we evaluate
(Veo-3.1, Kling-v3/Omni, Seedance-2.0, Hailuo-2.3, Vidu-Q3-Pro,
Grok-Imagine-Video and HappyHorse-1.x)---raised the ceiling to controllable,
audio-equipped clips, the community moved to disentangled, human-aligned
protocols. VBench~\citep{huang2024vbench} and VBench-2.0~\citep{huang2025vbench2}
decompose quality into 16 then 18 dimensions and push from \emph{superficial}
to \emph{intrinsic} faithfulness; Video-Bench~\citep{han2025videobench}
enlists MLLMs as scalable judges; VideoScore~\citep{he2024videoscore}
learns a human-aligned regressor. This stage established the grammar of
modern evaluation---quality as a structured set of interpretable
dimensions---but its axes are deliberately generic and its prompts are drawn
from web users, reflecting what is easy to source rather than what a film
production demands. The first gap is thus one of \emph{content and taxonomy}:
neither the prompts nor the dimensions are anchored in a professional
production grammar.

\paragraph{Specializing to emerging capabilities.}
With this foundation in place and baseline visual quality saturating,
benchmarks specialized to isolate the \emph{hard} capabilities that generic
scores had masked. Compositional suites
T2V-CompBench~\citep{sun2024t2vcompbench} and FETV~\citep{liu2023fetv} expose
brittle attribute, motion and spatial-relation binding; the temporal--motion
line ChronoMagic-Bench~\citep{yuan2024chronomagic} and
VMBench~\citep{ling2025vmbench} target metamorphic amplitude and
perception-aligned motion, while SLVMEval~\citep{slvmeval2026} extends
reliability testing to hour-scale footage. As generators began stitching
shots into a story, MSVBench~\citep{msvbench2026} introduced multi-shot
evaluation with a hybrid LMM-plus-expert scorer, distilling this stage's
lesson: current systems behave as \emph{visual interpolators rather than
world models}. A parallel reasoning wave then raised the bar from rendering
to understanding, probing structural (TiViBench~\citep{tivibench2026}),
social (SVBench~\citep{svbench2026}), broad multi-axis
(WorldJen~\citep{worldjen2026}) and embodied-robotic (RBench~\citep{rbench2026})
reasoning. Yet even MSVBench scores its shots against generic quality criteria
rather than a director's intended shot list, so evaluation still stops short of
the \emph{film}: cross-shot narrative structure and director intent are not
judged end-to-end against a verified cinematic reference. This is the second
gap.

\paragraph{Toward film: reference, audio, aesthetics and Cinematic Language.}
The strand closest to our goal relaxes that assumption toward
controllable, film-like generation, but does so one aspect at a time.
Reference/image-conditioned suites (ConsistI2V/I2V-Bench~\citep{ren2024consisti2v},
UI2V-Bench~\citep{ui2vbench2025} and IP-Bench~\citep{ipbench2026}) evaluate
consistency, semantic understanding and protection from a conditioning image,
raising \emph{reference fidelity} as a concern. As generators acquired joint
audio, AVGen-Bench~\citep{avgenbench2026} evaluates text-to-audio-video
generation, adding a \emph{cross-modal audio} axis, while
VGA-Bench~\citep{vgabench2026}, co-developed with the Beijing Film Academy,
brings aesthetic tagging into view. In parallel, a
cinematic-understanding literature (MovieNet, shot-type taxonomies,
CineScale/ShotBench) and storyboard-to-video pipelines treat film
as structured language, and MovieBench~\citep{wu2024moviebench} pushes to
feature-length identity consistency---sustained in at most $53\%$ of cases at
multi-scene scope. Collectively these efforts touch nearly
every ingredient of film, yet no single benchmark unifies them: aesthetics
and audio are not anchored in a production grammar, reference fidelity and
cross-shot audio--dialogue continuity are evaluated in isolation if at all,
the target is typically real or arbitrary footage rather than \emph{generated}
films, and the cinematic-understanding line asks only whether models can
\emph{read} Cinematic Language, never whether a generator can \emph{speak} it at
a director's granularity. Closing this third gap---a unified,
production-grounded evaluation of \emph{generated} film---is exactly the
remit of FilmBench.

\paragraph{Positioning of FilmBench.}
These three accumulated gaps (a generic taxonomy, a clip-level scope, and
fragmented coverage of individual film aspects) jointly define what a
film-grade benchmark must satisfy at once. To the best of our knowledge, no
existing benchmark simultaneously (i) reverse-engineers prompts from genuine,
award-winning films selected by directors; (ii) spans a broad set of
cinematic genres and both text-to-video and reference-to-video tasks, with
prompts organized as real, mostly multi-shot shot lists (1{,}056 of 1{,}169)
that evaluate the film rather than the isolated clip;
(iii) grounds its taxonomy in the Cinematic Language system taught in professional
film schools: 3 first-level axes, 12 second-level components and 35
third-level sub-metrics with an R2V-specific Visual-Following extension,
co-designed with directors, Beijing Film Academy faculty and a professional
film studio; and (iv) treats audio--dialogue continuity (including dialogue
intelligibility and per-character timbre stability across shots) as a
first-class axis. FilmBench thus asks not whether a model can \emph{read}
Cinematic Language but whether it can \emph{speak} it, and because every prompt is
reverse-engineered from a verified shot, we always retain both a structured
grammar and a ground-truth reference. \S\ref{sec:dim} and \S\ref{sec:eval}
detail this taxonomy and the automatic-plus-human evaluator stack, with a
head-to-head comparison table in \S\ref{sec:dim} and the appendix.

% !TeX root = neurips_2026.tex
% ===========================================================
% Section: Benchmark Construction and Dimensions
% ===========================================================
\section{FilmBench: Construction and Dimensions}
\label{sec:dim}

This section describes how FilmBench is built. We first present the
reverse-engineering pipeline (Figure~\ref{fig:construct_pipe}) that turns
genuine cinematic clips into
structured prompts (\S\ref{sec:dim:pipeline}), then introduce our
three-level Cinematic Language taxonomy of 3 first-level axes
(L1), 12 second-level components (L2) and 35 third-level sub-metrics (L3),
co-designed with faculty from the Beijing Film Academy and the Hujing
Digital Media~\& Entertainment Group film studio (\S\ref{sec:dim:taxonomy}).

\subsection{Reverse-Engineered Prompt Construction}
\label{sec:dim:pipeline}

FilmBench prompts are produced by a director-driven \emph{reverse-engineering}
pipeline with three stages (Figure~\ref{fig:construct_pipe}). \textbf{Clip
selection:} professional directors from the film studio (Hujing Digital
Media~\& Entertainment Group) and the Beijing Film Academy prioritize
\emph{award-winning films} (e.g., winners of the Academy Awards, the Golden
Horse Awards and the Hundred Flowers Awards) across our 20 cinematic genres,
and select \emph{multi-shot} clips (mostly $\ge 4$ shots)
that specifically stress Cinematic Language and visual-expression skill.
\textbf{Draft prompt inference:} a FilmOps operator suite together with a
strong multimodal model (Gemini~3.1~Pro) reverse-extracts the original
narrative script, Cinematic Language elements and on-screen visual elements from each
clip, which Gemini~3.1~Pro then assembles into an initial structured prompt.
\textbf{Expert refinement:} the Beijing Film Academy directing team screens,
quality-checks, rewrites and standardizes every draft into professional
film-grade Cinematic Language. Throughout clip selection and final curation,
the team ensures that
\emph{every L1 axis and L2 component in the taxonomy is covered}, so that the benchmark
faithfully probes a model's film-grade generation ability rather than
generic visual quality. Each prompt uses slotted cinematic tags
(\verb|@scene|, \verb|@role|, \verb|@prop|), e.g., ``Shot~1 (medium-shot,
eye-level fixed, rule-of-thirds): in a brightly lit ICU \verb|@scene1|, a
young female doctor \verb|@role1| $\dots$''.

\paragraph{Tasks and scale.}
FilmBench covers two tasks: \textbf{T2V} (prompt $\rightarrow$ video) and
\textbf{R2V} (reference image $+$ prompt $\rightarrow$ video). The T2V
suite contains \textbf{515 prompts} spanning \textbf{20 base cinematic
genres}, balanced across Chinese Films and International Films markets
(243 Chinese Films / 272 International Films). The R2V suite contains \textbf{654
prompts} split across three reference types (232 scenes / 209 props / 213
characters) and drawn from the same 20-genre pool across both markets
(239 Chinese Films / 415 International Films). Because prompts follow real
shot lists, most are \emph{multi-shot}: 402 of the 515 T2V prompts and all
654 R2V prompts script multiple shots (1{,}056 of 1{,}169 overall), unlike
the single-clip prompts of prior benchmarks. Each prompt
is reverse-engineered from a distinct award-winning film clip, so the prompt
count also reflects the number of source clips.

\subsection{Academy-Aligned Evaluation Taxonomy}
\label{sec:dim:taxonomy}

Our evaluation dimensions follow the academic Cinematic Language system taught in
professional film schools. The taxonomy is \emph{co-designed by faculty from
the Beijing Film Academy and the film studio of Hujing together with an analysis of
current video-generation models' capabilities}, and is deliberately
organized as a multi-level hierarchy so that conclusions and insights can be
drawn at coarse (L1), medium (L2) and fine (L3) granularity.

\begin{table}[!htbp]
\caption{FilmBench three-level taxonomy (3 L1 axes, 12 L2 components, 35 L3
sub-metrics). R2V adds a \emph{Visual Following} component (+3 L3). Full
definitions in Appendix~\ref{app:dims}.}
\label{tab:taxonomy}
\centering
\small
\begin{tabular}{l l}
\toprule
L1 Axis & L2 Component \\
\midrule
\multirow{5}{*}{Instruction Following (16+3)}
 & Cinematic Language \\
 & Character \& performance \\
 & Scene \\
 & Audio \\
 & \textit{Visual Following (R2V only)} \\
\midrule
\multirow{4}{*}{Temporal Continuity (7)}
 & Spatial coherence \\
 & Temporal coherence \\
 & Character coherence \\
 & Audio coherence \\
\midrule
\multirow{4}{*}{Aesthetic Quality (12)}
 & Base quality \\
 & Editing \\
 & Performance \\
 & Audio quality \\
\bottomrule
\end{tabular}
\end{table}

\paragraph{Three L1 axes.} FilmBench evaluates cinematic competence along
three first-level axes, decomposed into 12 L2 components and 35+3 L3
sub-metrics (Table~\ref{tab:taxonomy}); below we enumerate the Cinematic Language
sub-metrics and leave the remaining, more generic components to
Table~\ref{tab:taxonomy} and Appendix~\ref{app:dims}. \textbf{Instruction
Following} captures whether the video faithfully realizes the prompt's
cinematic intent, over four components: \emph{Cinematic Language} (shot scale,
camera movement, viewing angle, composition, focus, tone \& color, thematic
style), \emph{Character \& performance}, \emph{Scene}, and \emph{Audio}.
\textbf{Temporal Continuity} captures whether the output behaves as a film
rather than a single clip, over four components: \emph{spatial coherence},
\emph{temporal coherence}, \emph{character coherence}, and \emph{audio
coherence}. \textbf{Aesthetic Quality} captures production-grade quality, over
four components: \emph{base quality}, \emph{editing} (editing fluency,
camera-work appeal), \emph{performance} (emotional performance, action
performance), and \emph{audio quality}.

\paragraph{Task-specific extension.} T2V and R2V share this main framework.
R2V additionally introduces one L2 component, \textbf{Visual Following}, with
three L3 sub-metrics (\emph{scene-space}, \emph{character-appearance} and
\emph{prop-reference} fidelity), each measuring adherence to the conditioning
reference image. Every R2V prompt conditions on one of three reference types
(\emph{scene}, \emph{prop} or \emph{character}), and only the matching
fidelity sub-metric is scored for that prompt. Visual Following is scored
only for R2V, so R2V has 13
L2 and 35+3 L3 dimensions, while all other dimensions are identical across tasks.
For every L3 sub-metric we provide a 5-point anchor rubric; the complete
definition table is in Appendix~\ref{app:dims}.

% !TeX root = neurips_2026.tex
% Section: Evaluation Method
\section{Evaluation Method}
\label{sec:eval}

FilmBench is scored by an expert-grade automatic evaluation agent which integrates a suite
of Cinematic Language \emph{operators}, expert annotation models, and a judge
model. For every dimension in the taxonomy the agent produces a 1--5 score, achieving high agreement with human experts
(\S\ref{sec:eval:agg}). Below we describe the open-sourced operator suite
(\S\ref{sec:eval:ops}) that grounds the Cinematic Language dimensions and
the score-aggregation rule (\S\ref{sec:eval:agg}).

\subsection{FilmOps: A Cinematic Language Operator Suite}
\label{sec:eval:ops}

A key obstacle to professional cinematic evaluation is that neither generic
MLLM-as-judge approaches nor existing domain expert models reliably encode
film-industry visual rules: general multimodal models are trained on
open-domain understanding and misjudge professional categories such as shot
scale, composition and camera movement~\citep{wang2025cinetechbench}, while existing aesthetic/shot models
are trained mostly on everyday photography or live-action footage and do not
transfer across genres (3D/2D animation, stylized content). We therefore
build \textbf{FilmOps}, an open-source operator suite that maps a generated
video into structured cinematic labels for the Cinematic Language
dimensions.

\paragraph{Industry-aligned taxonomy.} FilmOps defines a professional
classification system, aligned with classical production references and
vetted by front-line practitioners, over six core dimensions:
\emph{shot scale}, \emph{composition}, \emph{viewing angle}, \emph{tone \&
color}, \emph{character layout}, and \emph{camera movement}. Five are
frame-level and one (camera movement) is shot-level; together they span 55
sub-categories (character layout is an open-ended natural-language field).
Categories are defined by narrative/visual function rather than imaging
mechanism, so the same standard applies across live-action, 3D and 2D
genres.

\paragraph{Multi-genre training data.} To match the cross-genre, multi-style
distribution of generated video, FilmOps is trained on a pool of 5{,}000+
real film/TV works spanning live-action, 3D animation, 2D animation and
stylized/VFX content, and across narrative types (dialogue-driven, action,
group scenes, and long-tail shot forms). Each operator is trained on
$\sim$40K--60K annotated samples with a shot-based test set, under a strict
annotator-training and quality-control protocol led by professional
practitioners.

\paragraph{Task-matched operator design.} Operators are built to match each
dimension's characteristics: frame-level visual
dimensions (shot scale, composition, angle, color) use vision backbones
(DINO, BEiT and ResNet-18), while dimensions requiring relational
reasoning or temporal modeling (character layout, camera movement) use a
multimodal model (InternVL3-14B, LoRA-SFT). Classification operators are
evaluated by macro-F1 and the natural-language layout operator by precision.
Against four strong zero-shot general-MLLM baselines, the trained operators
improve markedly on every dimension, confirming the value of
domain-specialized operators; the full per-operator backbones and macro-F1 figures tested over 400 images/shots against these baselines are reported in the appendix
(Table~\ref{tab:operators}).

To support customizable, personalized self-evaluation, we \textbf{open-source
the classification standard, model weights and inference scripts} of FilmOps.
The modular design is not tied to a single model, so users can reuse
individual operators or embed specific dimensions into their own evaluation
pipelines.

\subsection{Score Aggregation}
\label{sec:eval:agg}

Each L3 sub-metric receives a 1--5 score, which is \textbf{linearly mapped to
a 0--100 scale before aggregation} ($1\!\to\!0, 2\!\to\!25, 3\!\to\!50,
4\!\to\!75, 5\!\to\!100$). Aggregation is \textbf{sample-level}: each video
is first aggregated across its dimensions, and per-model scores are the mean
over that model's videos, preserving sample variance for significance
analysis. We aggregate \textbf{L3~$\rightarrow$~L1 directly} (bypassing L2), treating every L3 sub-metric as equally important; each L1 axis
is the mean of its constituent L3 sub-metrics, and the Overall score is the
equal-weighted mean of the three L1 axes; L2 components are computed as a
side branch for fine-grained reporting only. Not-applicable scores
(N/A or null) are excluded from aggregation; for models without audio capability,
audio dimensions are handled symmetrically as not-applicable.

\begin{figure}[H]
\centering
\includegraphics[width=\linewidth]{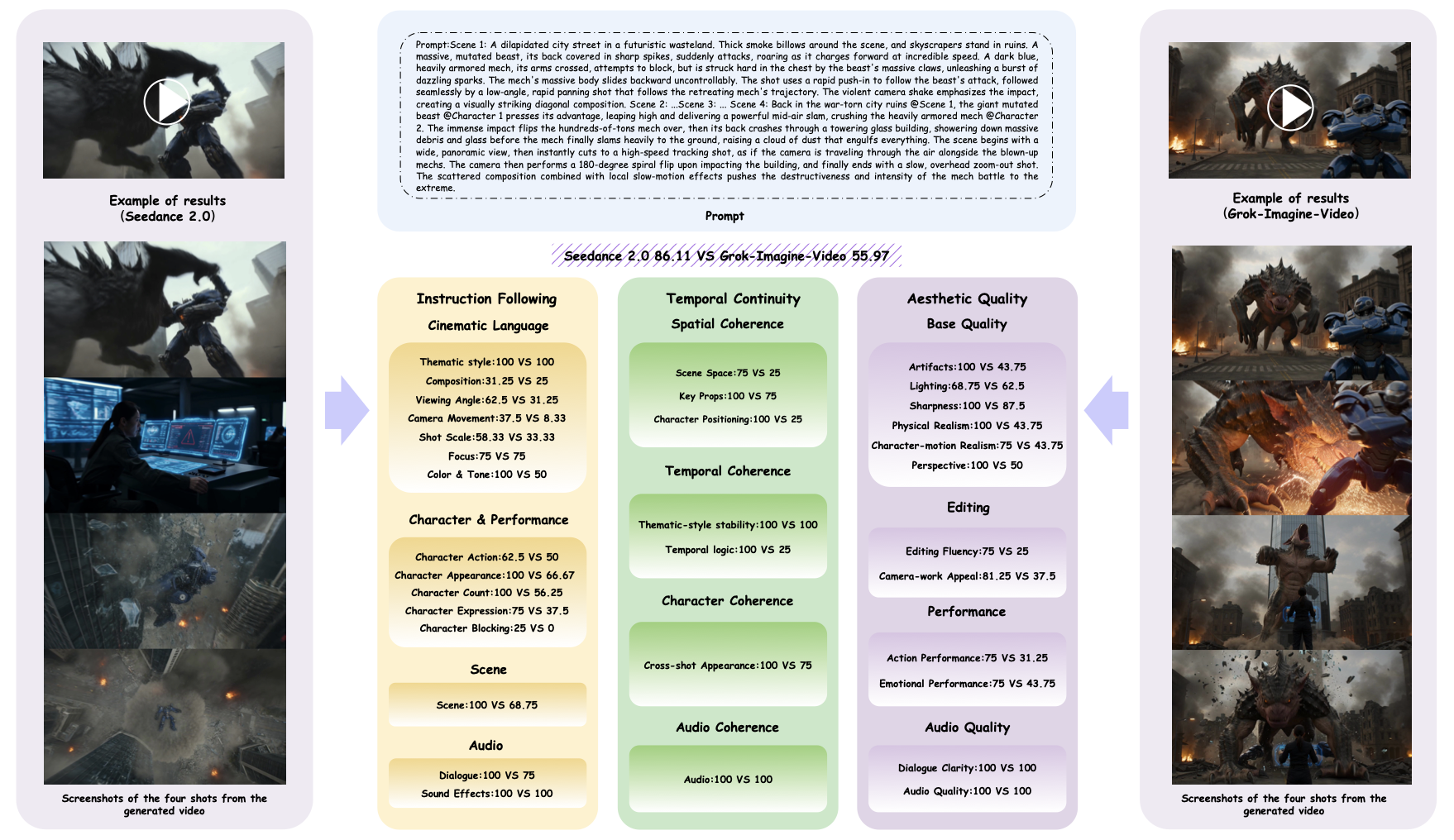}
\caption{Multi-shot T2V example (a sci-fi mech battle): Seedance~2.0
(86.11) vs.\ Grok~Imagine~Video (55.97). Discussed in the text.}
\label{fig:eval_example_4}
\end{figure}

\paragraph{Qualitative case studies.}
Fine-grained per-dimension scoring exposes trade-offs that aggregate scores
hide. As a worked example, Figure~\ref{fig:eval_example_4} scores a multi-shot
T2V sci-fi mech battle, a case that stresses the dual challenge of intense
action combined with multi-shot cutting. Seedance~2.0 (86.11) clearly
outperforms Grok~Imagine~Video (55.97), and the gap is diagnostic rather than
uniform: Seedance is markedly stronger on tone \& color, character
\&~performance and scene, and on camera-work appeal; in the action passages
its character positioning stays consistent across shots and its temporal logic
shows no visible errors. Grok, by contrast, exhibits scene drift and drifting
cross-shot appearance and positioning, which surfaces directly in the
per-dimension gaps on character-motion realism and action performance. The
introduction (Figure~\ref{fig:eval_example}) shows a complementary multi-shot
R2V reference-scene case, and Appendix~\ref{app:examples} provides two further
worked examples (reference-character and reference-prop) that together cover
the major scoring scenarios in FilmBench.

% !TeX root = neurips_2026.tex
% ===========================================================
% Section: Experiments  (真实结果, stand 带站位版)
% !! 已更新至 20260713_v2 stand (剔除版): T2V 515 prompts, R2V 654 prompts / agreement overlap(交集) T2V 120 / R2V 221 !!
% 数值来源: results_analyze/Score_L3231_Mapped_stand{,_r2v}_20260713_v2/
% 图: figures/*_t2v.pdf 与 figures/*_r2v.pdf (英文版, 由 draw_figures_en.py 生成)
% ===========================================================
\section{Experiments}
\label{sec:exp}

\subsection{Models and Setup}
\label{sec:exp:setup}

We benchmark leading video-generation models under each task. For
\textbf{T2V} we evaluate 9 models: Seedance~2.0~\citep{seedance2026seedance20}, HappyHorse~1.1~\citep{happyhorse2026v11},
HappyHorse~1.0~\citep{happyhorse2026v10}, Kling~3.0~\citep{klingteam2025klingomni}, Kling~3.0~Omni~\citep{klingteam2025klingomni}, Veo~3.1~\citep{google2025veo3,wiedemer2025video}, Grok~Imagine~Video~\citep{xai2026grok},
Vidu~Q3~Pro~\citep{bao2024vidu} and Hailuo~2.3~\citep{minimax2025hailuo23}. For \textbf{R2V} we evaluate 7 models: Seedance~2.0, HappyHorse~1.1,
HappyHorse~1.0, Kling~3.0~Omni, Veo~3.1, Grok~Imagine~Video,
and Vidu~Q2~Pro~\citep{bao2024vidu}.
All models are run under their official APIs / public
checkpoints with default settings, generating one video per prompt. This
yields $9\times515=4{,}635$ machine-scored T2V records and
$7\times654=4{,}578$ R2V records. Throughout, T2V and R2V are reported with
the \emph{same} analysis format: overall, per-axis (L1), per-component
(L2), per-sub-metric (L3), and discriminability, so
that the two tasks are directly comparable; R2V additionally carries a
reference-type analysis (scene / character / prop).

\subsection{Human Agreement}
\label{sec:exp:agree}

We validate the automatic evaluator against 90 professional human raters
majoring in film, television, directing, or digital media from the
Beijing Film Academy at the \emph{model level}: over 300 prompts
($\sim$30\% of the benchmark) carry both machine and human judgments,
yielding $2{,}878$ model--prompt pairs. We compute per-model machine and
human means at each L2 component and correlate the two rankings
(Table~\ref{tab:agree}). The overall model-level Spearman $\rho$
reaches $0.95$ (T2V) and $0.96$ (R2V), confirming that the automatic
leaderboard closely reproduces expert model rankings.

\begin{table}[!htbp]
\caption{L2 component-level human agreement (Spearman $\rho$), averaged
across T2V and R2V. 10 of 13 components reach $\rho \geq 0.78$;
the three lower-agreement components (audio quality, audio coherence,
editing) are separated by a mid-rule.}
\label{tab:agree}
\centering
\resizebox{\linewidth}{!}{%
\begin{tabular}{l*{13}{c}}
\toprule
 & \rotatebox{55}{\small Perform.} & \rotatebox{55}{\small Vis.\ Follow.} & \rotatebox{55}{\small Cinem.\ Lang.} & \rotatebox{55}{\small Base qual.} & \rotatebox{55}{\small Audio} & \rotatebox{55}{\small Char.\ \& perf.} & \rotatebox{55}{\small Temp.\ coh.} & \rotatebox{55}{\small Spat.\ coh.} & \rotatebox{55}{\small Scene} & \rotatebox{55}{\small Char.\ coh.} & \rotatebox{55}{\small Audio q.} & \rotatebox{55}{\small Audio coh.} & \rotatebox{55}{\small Editing} \\
\midrule
$\rho$ & 0.99 & 0.96 & 0.95 & 0.95 & 0.93 & 0.92 & 0.85 & 0.83 & 0.82 & 0.78 & 0.68 & 0.68 & 0.60 \\
\bottomrule
\end{tabular}}
\end{table}

Drilling into the component level, 10 of the 13~L2 dimensions achieve
$\rho \geq 0.78$; the three lower-agreement components---\emph{audio
quality} ($0.68$), \emph{audio coherence} ($0.68$) and \emph{editing}
($0.60$)---are those where subjective temporal behaviour (sound
fidelity, cut rhythm) is hardest to pin down, and we acknowledge that
human and automatic judgments diverge more on these axes.  The
overall-level agreement ($\rho \geq 0.95$ on both tasks) nonetheless
indicates that the aggregate ranking is a reliable proxy for expert
judgment, and we flag the three lower-agreement dimensions when
interpreting fine-grained results.

\subsection{Overall Leaderboard}
\label{sec:exp:main}

\begin{figure}[!htbp]
\centering
\includegraphics[width=0.86\linewidth]{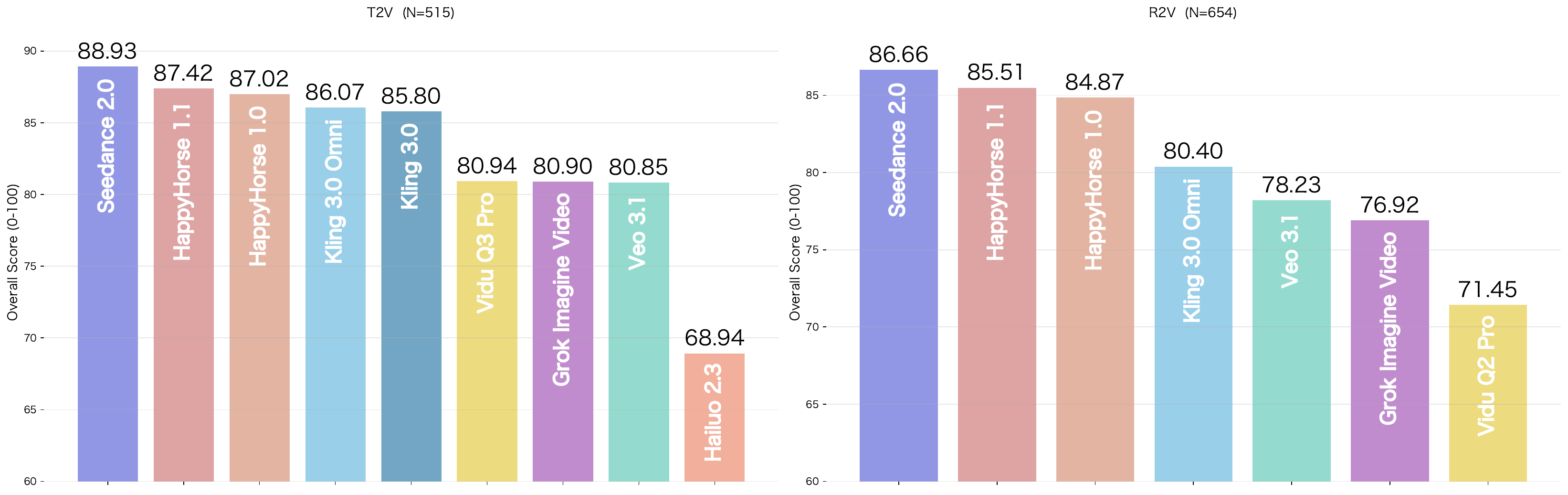}
\caption{Overall FilmBench scores (0--100) per model; model names are printed
inside each bar and the dashed line is the cross-model average. Left: T2V
($N=515$); right: R2V ($N=654$).}
\label{fig:overall}
\end{figure}

Tables~\ref{tab:t2v-pillar}--\ref{tab:r2v-pillar} and Figure~\ref{fig:overall} report the overall
FilmBench score (0--100, higher is better). On T2V, Seedance~2.0 leads
(88.93), closely followed by HappyHorse~1.1 (87.42) and HappyHorse~1.0
(87.02); the two Kling~3.0 variants form a second tier ($\sim$86), Vidu~Q3~Pro,
Grok and Veo~3.1 a third tier ($\sim$81), with Hailuo~2.3 at
68.94. On R2V, the top group is Seedance~2.0 (86.66), HappyHorse~1.1 (85.51)
and HappyHorse~1.0 (84.87). Crucially, no
model approaches saturation, in sharp contrast with the near-ceiling scores
reported on prior web-style benchmarks, and the top group is consistent
across both tasks, indicating that reference conditioning does not reshuffle
the leaders.

\subsection{Dimension Variance}
\label{sec:exp:var}

\begin{figure}[!htbp]
\centering
\includegraphics[width=\linewidth]{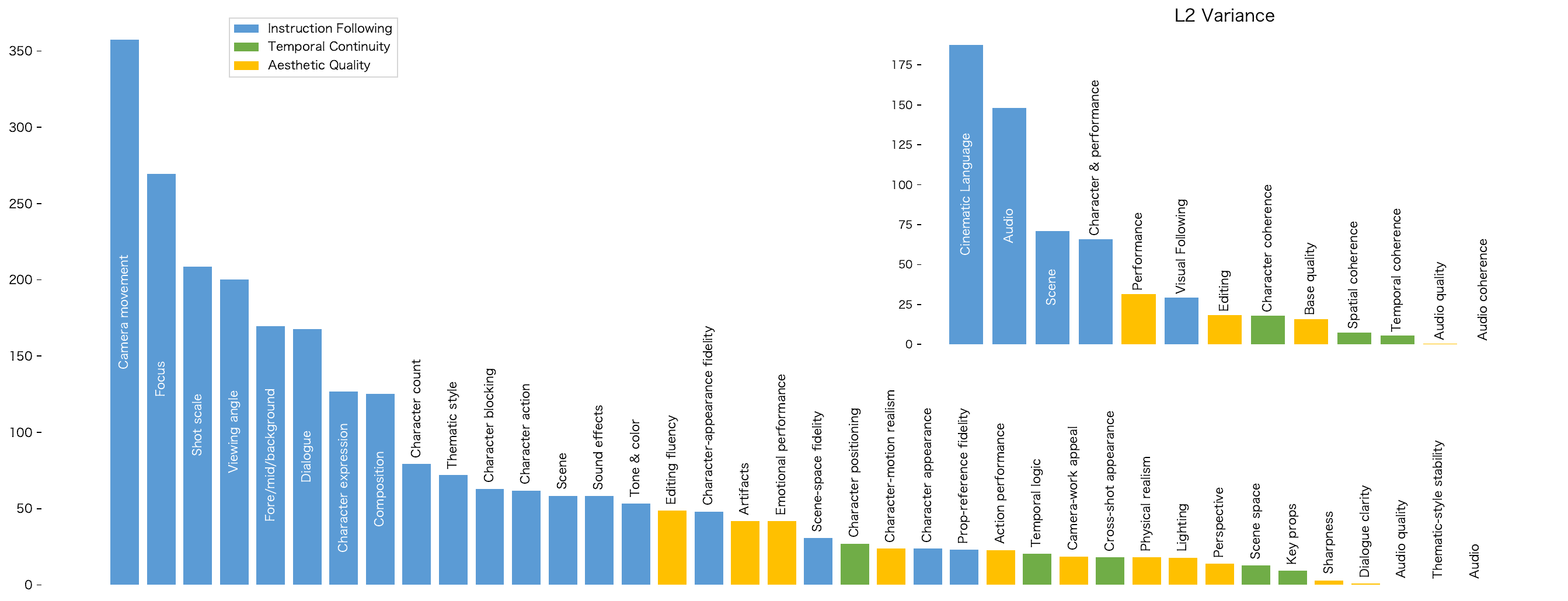}
\caption{Cross-model variance per L3 sub-metric (main bars, colored by
axis), with L2 inset (upper right), computed over the merged T2V+R2V model
means (Hailuo excluded).}
\label{fig:var_combined}
\end{figure}

To quantify where models differ most at fine-grained levels, we compute the
cross-model variance of each dimension across the hierarchy
(Figure~\ref{fig:var_combined}; Hailuo is excluded from the
variance computation to avoid its audio outliers dominating).
At the axis level, \emph{Instruction Following} shows by far the
largest variance ($99.4$), nearly eight times that of Aesthetic Quality
($12.6$) and seventeen times that of Temporal Continuity ($5.6$),
indicating that models differ primarily in their ability to execute
professional cinematic language instructions.

Drilling into the component level, \emph{Cinematic Language} dominates with
a variance of $187.9$, far exceeding Audio ($148.4$), Scene ($71.3$)
and all other components.  At the sub-metric level, the top five
highest-variance dimensions are all Cinematic Language or scene-related:
camera movement ($357.6$), focus ($269.6$), shot scale ($209.0$),
viewing angle ($200.5$) and fore/mid/background ($169.8$).
This concentration confirms that the primary differentiator among
current generative video models lies in Cinematic Language instruction
following---whether a model understands and executes professional
camera-language conventions.
Per-task breakdowns are provided in Appendix~\ref{app:variance}.

\subsection{Per-Axis Landscape (L1)}
\label{sec:exp:pillar}

\begin{figure}[!htbp]
\centering
\includegraphics[width=\linewidth]{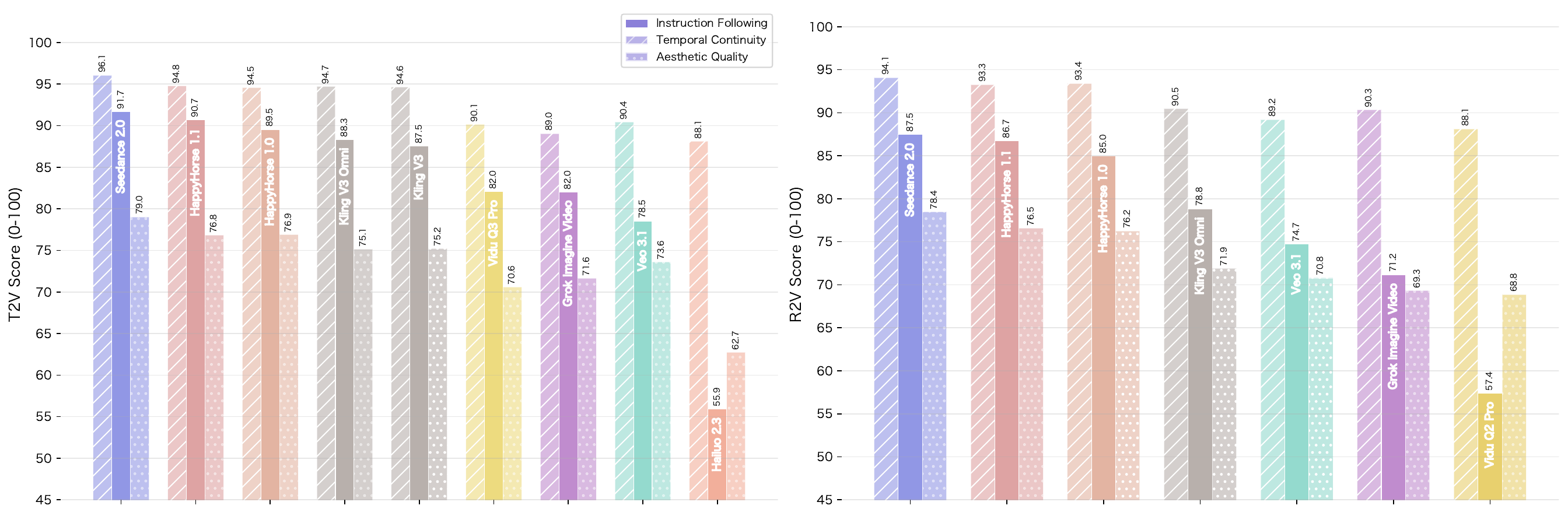}
\caption{Per-axis rankings for T2V (left, 9 models) and R2V (right, 7 models).
Each model is scored on three L1 axes: Instruction Following (solid),
Temporal Continuity (hatched), and Aesthetic Quality (dotted). Models are
ranked by overall score; R2V Instruction Following includes the visual-following
component.}
\label{fig:pillar_combined}
\end{figure}

Across the three axes, models are most differentiated on
\emph{Instruction Following}, which encompasses the Cinematic Language
sub-metrics that drive the high variance reported above. On T2V
(Figure~\ref{fig:pillar_combined}, left), Seedance~2.0 leads all three axes:
Instruction Following (91.69, ahead of HappyHorse~1.1 at 90.70), Temporal
Continuity (96.06) and Aesthetic Quality (79.03). On R2V
(Figure~\ref{fig:pillar_combined}, right), Seedance~2.0 likewise leads every axis
(Instruction Following 87.45, Temporal Continuity 94.09, Aesthetic Quality
78.44). The Instruction Following axis shows the widest inter-model
gap under both tasks, as the Cinematic Language and scene-related
sub-metrics amplify the spread in professional-language execution.

\subsection{Per-Component Rankings (L2)}
\label{sec:exp:l2}

\begin{figure}[!htbp]
\centering
\includegraphics[width=\linewidth]{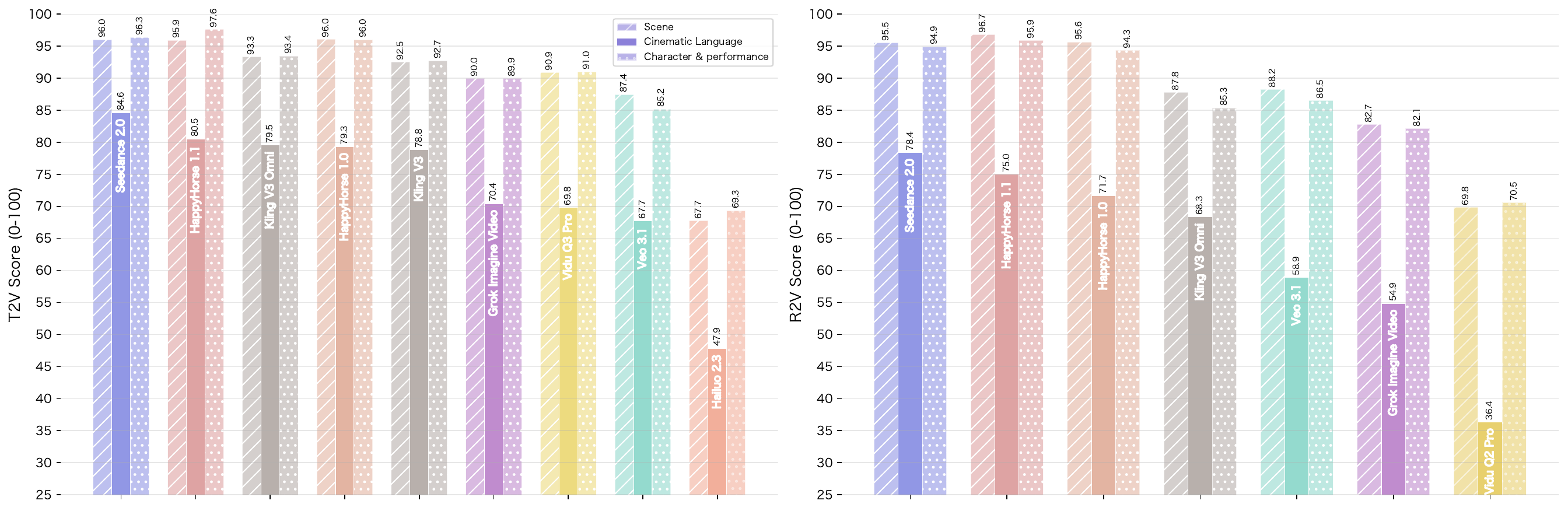}
\caption{Per-model rankings within the L2 components most related to
audiovisual language: \emph{Cinematic Language} (solid, centre, model names
inside; also the sort key), flanked by \emph{Scene} (hatched) and
\emph{Character~\&~performance} (dotted). T2V (left, 9 models) and R2V
(right, 7 models). The full L2 rankings are in the appendix.}
\label{fig:l2top3}
\end{figure}

Descending one level, we rank models within the L2 components most closely
tied to \emph{Cinematic Language} together with the
adjacent \emph{Scene} and \emph{Character~\&~performance} components
(Figure~\ref{fig:l2top3}); the full 12/13-panel rankings are deferred to the
appendix. Cinematic Language is placed at the centre and used as the sort key,
and each component isolates a different craft skill, revealing where the head
models specialize:

\textbf{Cinematic Language} (camera work, framing, transitions) is the sharpest
discriminator of the three, with the field fanning out over a 17--42-point
range and even the strongest model reaching only 84.6. This is the one
component that hinges on \emph{temporal} decision-making---when to move the
camera, how to reframe, when to cut---rather than per-frame rendering, which
is why models diverge so widely here. Seedance~2.0 leads decisively and is the
only model whose camera competence holds up when a reference image is imposed,
marking dynamic camera control as its signature strength.

\textbf{Scene} (environment consistency, lighting, background layout) also
spreads the field, though the leading cluster reaches the mid-90s
(HappyHorse~1.0 at 96.0, HappyHorse~1.1 at 96.7). The gap opens further down
the ranking, where weaker models struggle to keep a coherent, consistently
lit environment across shots. The separation here is owned by the HappyHorse
family, whose strength lies specifically in preserving spatial layout---a
capability distinct from camera craft, since the models that lead scene
rendering are not the ones that lead Cinematic Language.

\textbf{Character~\&~performance} spreads the field the least of the three but
still meaningfully, with the leaders reaching the high-90s (HappyHorse~1.1 at
97.6, Seedance~2.0 at 96.3). Here differentiation comes from sustaining
believable characters and acting across a full scene, and HappyHorse~1.1 holds
a slim but consistent edge, making it the character specialist just as
Seedance~2.0 is the camera specialist. Read together, the three components
show that the overall leaderboard masks a division
of labor---camera craft, scene rendering and character performance are led by
different models---so no single system dominates every axis of film craft.

\subsection{Fine-Grained Rankings (L3)}
\label{sec:exp:l3}

\begin{figure}[!htbp]
\centering
\includegraphics[width=\linewidth]{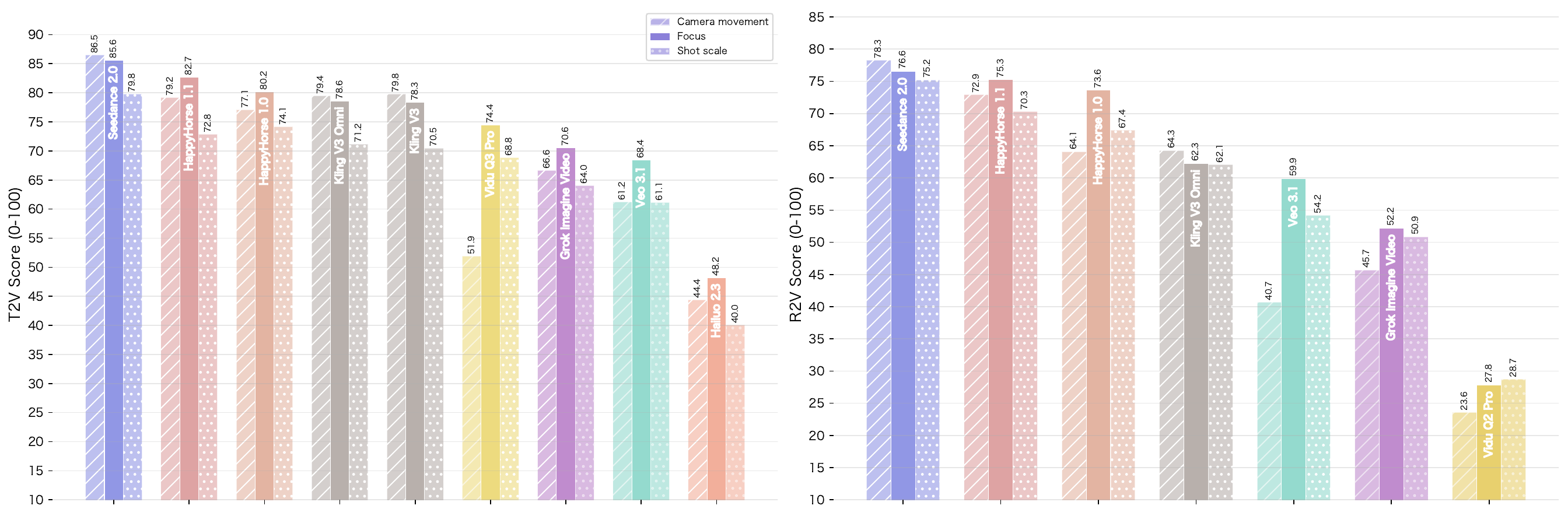}
\caption{Per-model rankings for the three highest-variance L3 sub-metrics
(computed over merged T2V+R2V data): \emph{Camera movement} (hatched),
\emph{Focus} (solid, model names inside), and \emph{Shot scale} (dotted).
T2V (left, 9 models) and R2V (right, 7 models).
The full per-metric panels are in the appendix.}
\label{fig:l3top3}
\end{figure}

At the finest level, FilmBench resolves 35 (T2V) / 38 (R2V) L3 sub-metrics,
each with its own ranking; Figure~\ref{fig:l3top3} shows the three sub-metrics
selected by highest merged T2V+R2V variance. All three are
Cinematic Language sub-metrics, indicating that \textbf{camera craft is where the
leading models pull apart most}:

\textbf{Camera movement} shows the largest spread among the head models, with
scores ranging from Seedance~2.0's 86.5 down toward the low 50s. Executing a
motivated camera move---a push-in, a pan that tracks action, a crane
reveal---requires the model to sustain a coherent scene through time, and
models diverge considerably in how convincingly they do so. Seedance~2.0
leads by a clear margin, suggesting a training emphasis on dynamic camera
choreography that sets it apart from the rest of the field.

\textbf{Focus} (depth-of-field control, subject isolation) sees the leaders
cluster in the low-to-mid 80s (Seedance~2.0 at 85.6, HappyHorse~1.1 at 82.7),
with a wider tail below. The metric separates models because it rewards
\emph{intentional} focus that directs attention to the dramatically relevant
subject---a semantic judgement, beyond merely producing a shallow-depth
look, that the stronger models handle more reliably.

\textbf{Shot scale} (close-up, medium, wide framing decisions) tests whether a
model chooses the framing a scene calls for---an intimate close-up for a line
of dialogue, a wide for an establishing beat. Seedance~2.0 again leads (79.8)
with HappyHorse~1.1 the runner-up, and the head ordering across all three
sub-metrics is consistent---Seedance~2.0 first, the HappyHorse family close
behind. Because these are the dimensions on which models differ most, holding
a high standard across them is what distinguishes a film-grade leader:
Seedance~2.0's overall lead comes not from a single capability but from
staying near the top on the most challenging Cinematic Language dimensions at
once, while the HappyHorse family remains competitive across them.

Crucially, \textbf{no model wins on all L3 sub-metrics}:
Seedance~2.0, the overall leader, claims only 18/35 (T2V) or 20/38 (R2V)
championships, concentrated on Cinematic Language sub-metrics---Cinematic Language
(camera movement, shot scale, focus, viewing angle), camera-work appeal, and
performance (action and emotional performance)---in both tasks.
HappyHorse~1.1, the runner-up, dominates 11/35 (T2V) or 10/38 (R2V)
sub-metrics on character instruction-following, audio, and scene dimensions.
Notably, on the three R2V-specific visual-following sub-metrics (scene-space
fidelity, character-appearance fidelity and prop-reference fidelity), the
HappyHorse family takes all three championships---HappyHorse~1.0 leads
scene-space and character-appearance fidelity, HappyHorse~1.1 leads
prop-reference fidelity. This ``championship mismatch'' is the key
value of fine-grained evaluation: a single leaderboard number conceals the
structural complementarity between models. Beyond the top two, the remaining
championships are scattered across several models, with 4~(T2V) or 2~(R2V)
models winning no championships at all. This granularity turns a single leaderboard into an actionable
diagnostic: a mid-tier model can still lead on individual craft dimensions.
The complete per-model score matrices (L3 and L2 heatmaps) are in the appendix
(Figures~\ref{fig:heat_l3_t2v}--\ref{fig:heat_l2_r2v}).

\subsection{Market and Genre Robustness}
\label{sec:exp:market}

\begin{figure}[!htbp]
\centering
\includegraphics[width=\linewidth]{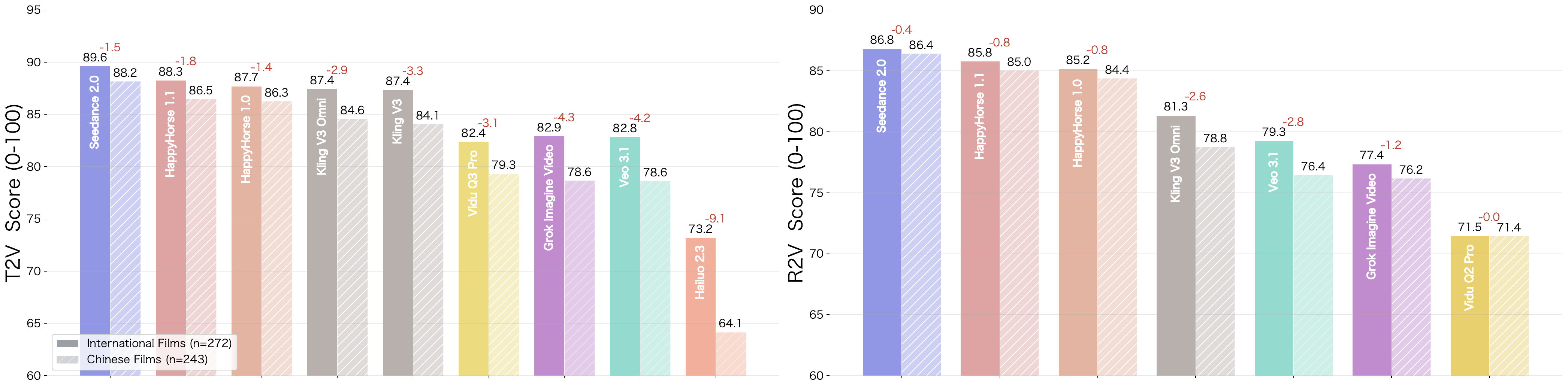}
\caption{Overall ranking split by market, Chinese Films vs.\ International Films;
left: T2V ($n=243$/$272$), right: R2V ($n=239$/$415$).}
\label{fig:market}
\end{figure}

\begin{figure}[!htbp]
\centering
\includegraphics[width=0.92\linewidth]{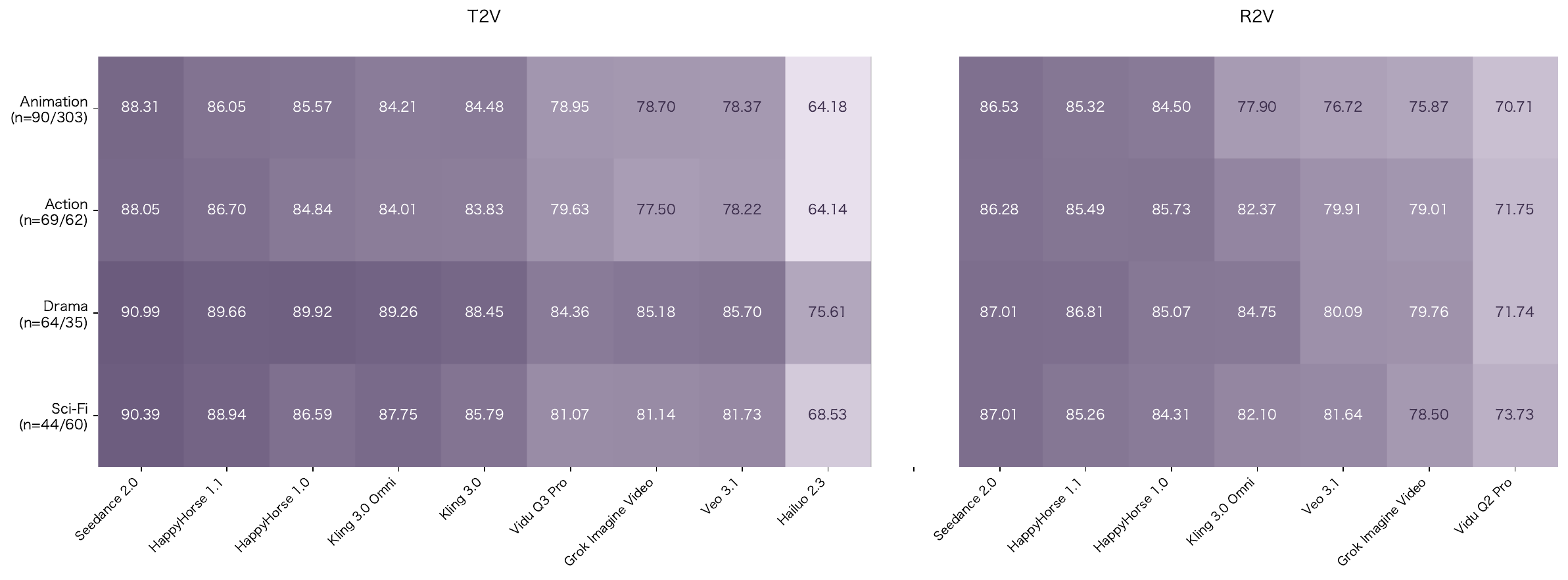}
\caption{Overall score heatmap across the top 4 genres by sample size. Y-axis: genres (with sample sizes T2V/R2V); X-axis: models split into T2V (left) and R2V (right) sections. Each cell shows the genre-specific mean score.}
\label{fig:genre}
\end{figure}

FilmBench spans 20 cinematic genres across Chinese Films and International Films
markets (T2V $n=243$/$272$; R2V $n=239$/$415$) to ensure broad diversity and
coverage.
Figure~\ref{fig:market} splits each task by market: the top group
(Seedance~2.0, HappyHorse~1.1/1.0) is identical for Chinese Films and
International Films prompts on both tasks, with only minor mid-pack reordering. Figure~\ref{fig:genre} shows the score heatmap across the top 4 genres by
sample size: rankings remain stable across genres, and each model's per-genre
spread is small ($\pm$2--3 points), confirming that both leaderboards reflect
model capability rather than genre mix. The action subset, often cited as the hardest, shows
the same ranking with only a slightly lower ceiling.

\subsection{Prompt Complexity and Content Factors}
\label{sec:exp:content}

Beyond market and genre, we probe three prompt-level factors that stress
generation differently: narrative structure (single- vs multi-shot prompts),
content type (action vs dialogue scenes), and rendering style (live-action vs
animation). In T2V, 113 of 515 prompts are single-shot and 402 are multi-shot;
every R2V prompt is multi-shot, so the shot contrast is T2V only. We tag 117 of
515 T2V prompts and 190 of 654 R2V prompts as action, and label all 654 R2V
prompts as live-action (486) or animation (168).

\begin{figure}[!htbp]
\centering
\includegraphics[width=\linewidth]{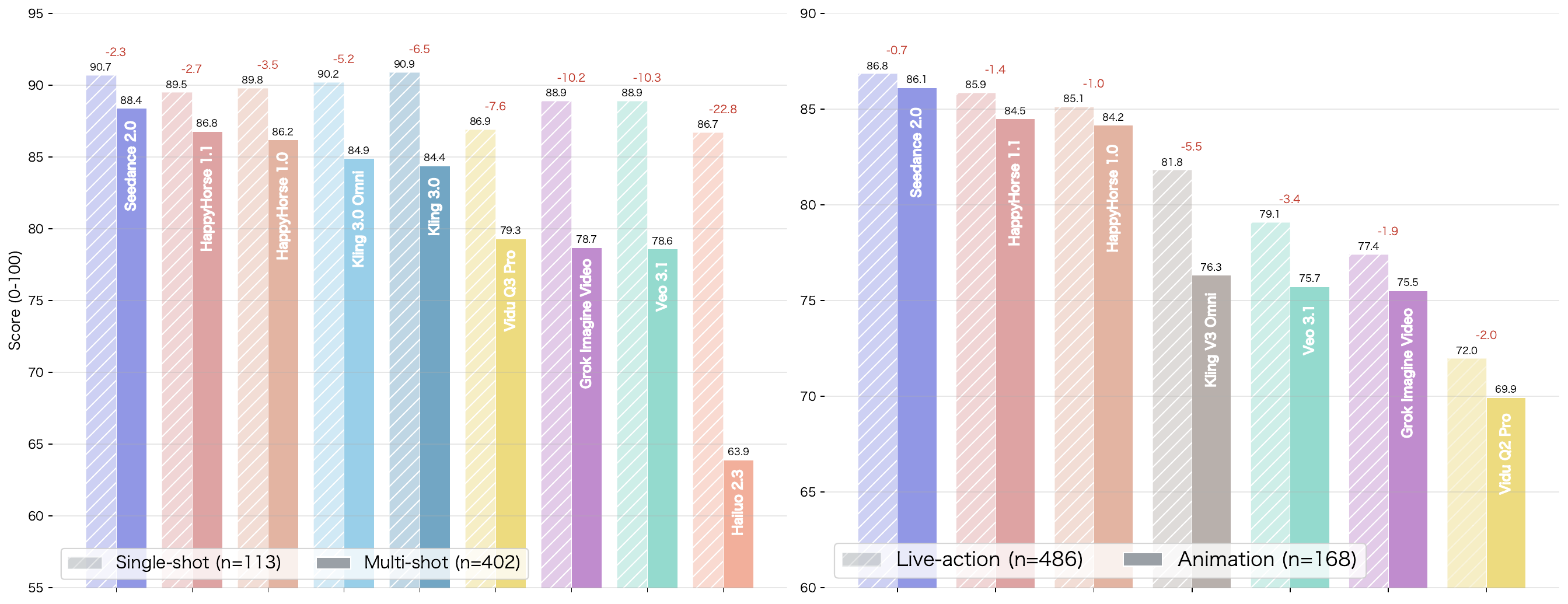}
\caption{Left: T2V single-shot (hatched) vs multi-shot (solid) mean scores (0--100)
per model, ordered by multi-shot score; red numbers give the drop $\Delta$;
$n{=}113$/$402$. Right: R2V live-action (hatched) vs animation (solid) mean
scores per model; $n{=}486$/$168$.}
\label{fig:shot-style}
\end{figure}

\textbf{Multi-shot is uniformly harder.} Every model scores lower on multi-shot
prompts (Figure~\ref{fig:shot-style}, left), with an average drop of 7.9 points (89.2 to 81.3).
The drop is larger for lower-ranked models: the top models lose little
(Seedance~2.0 $-2.3$, HappyHorse~1.1 $-2.7$), whereas lower-ranked models show
much larger drops (Grok~Imagine~Video $-10.2$, Veo~3.1 $-10.3$,
Hailuo~2.3 $-22.8$). The degradation concentrates on shot-craft dimensions
(Table~\ref{tab:shot_dims} in the appendix): Cinematic Language (composition, viewing angle, camera movement, focus),
editing fluency, and cross-shot spatial consistency, while audio and
temporal-coherence dimensions move little. A single-shot prompt only tests
intra-shot rendering, whereas a multi-shot prompt additionally requires planning
distinct shots and cutting between them inside one clip; the wider inter-model
spread on multi-shot prompts reflects a compositional competence that current
generators have not yet mastered, and we therefore treat shot structure as a
first-class evaluation axis rather than folding it into a single leaderboard
number.

\begin{figure}[!htbp]
\centering
\includegraphics[width=\linewidth]{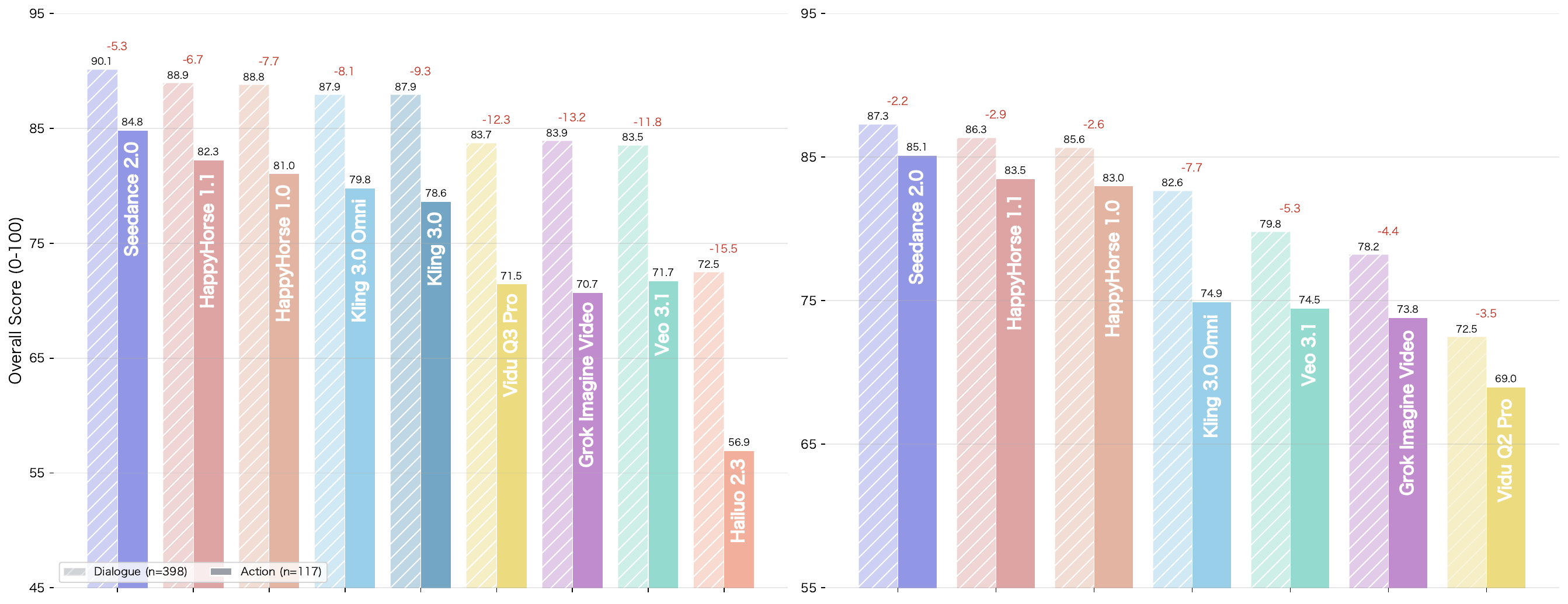}
\caption{T2V (left, 9 models) and R2V (right, 7 models): dialogue (hatched)
vs.\ action (solid) mean scores (0--100) per model, ordered by dialogue score;
red numbers give the drop $\Delta$ (action $-$ dialogue). T2V: action
$n{=}117$, dialogue $n{=}398$. R2V: action $n{=}190$, dialogue $n{=}464$.}
\label{fig:action}
\end{figure}

\textbf{Action scenes uniformly lower scores across all models.} Every model
scores lower on action than on dialogue prompts
(Figure~\ref{fig:action}), and, as with multi-shot prompts, the drop is larger
for lower-ranked models: the top group loses least (T2V Seedance~2.0 $-5.3$,
HappyHorse~1.1 $-6.7$) while lower-ranked models show larger drops (Vidu~Q3~Pro
$-12.3$, Hailuo~2.3 $-15.5$). \textbf{Seedance~2.0 maintains the lead on
action-specific L3 sub-metrics:} on T2V action prompts it tops 16/35
sub-metrics, notably character action (99.3), character expression (96.7),
fore/mid/background (95.0) and viewing angle (91.1); on R2V action
prompts it tops 11/38 sub-metrics, notably tone \& color (100.0), camera
movement (84.9), composition (83.1) and shot scale (78.2). Its action drops
on these dimensions remain moderate (T2V $\leq$3.6, R2V $\leq$10.7),
indicating that its Cinematic Language and character-performance strengths carry
over to action content. \textbf{The action performance degradation concentrates in
Instruction Following and Aesthetic Quality, with two distinct L3 clusters}
(Table~\ref{tab:action_dims} in the appendix). In Instruction Following the drop is dominated
by camera-work sub-metrics---camera movement ($-31.1$), focus ($-28.6$), shot
scale ($-26.6$) and composition ($-21.0$)---indicating that models show
larger drops in camera framing and motion once the scene becomes
fast and multi-agent. In Aesthetic Quality a second cluster of
\emph{physical-realism} sub-metrics degrades sharply: physical plausibility
($-16.9$), character-motion realism ($-15.8$), perspective ($-15.3$), artifacts
($-11.9$) and clarity ($-11.8$). Action scenes thus reveal two concurrent
degradation patterns that an aggregate score hides---models not only shift
camera framing, but also show reduced physical plausibility and increased
artifacts under vigorous
motion. The same two clusters lead the R2V action drop, at a smaller magnitude
(physical plausibility $-9.9$, artifacts $-3.5$); we report this contrast
descriptively, as the two tasks differ in prompt and sample distribution.

\textbf{Top models already generalize across visual style.} On R2V, the
leading group scores almost identically on live-action and animation
(Figure~\ref{fig:shot-style}, right; Seedance~2.0 $86.8$ vs.\ $86.1$, gap $<1.5$),
indicating that rendering style is no longer a constraint at the frontier.
Style generalization instead shows wider variation in the mid-tier: several mid-ranked models
score markedly lower on animation than live-action (Kling~3.0~Omni
$81.8\!\to\!76.3$, $-5.5$), and the loss again concentrates in Instruction
Following (Kling~3.0~Omni $80.9\!\to\!72.8$) and Aesthetic Quality, echoing the
action analysis: the two harder distributions, action and animation, both
challenge models on the same two axes.

\subsection{R2V: Reference Types}
\label{sec:exp:r2v}

\begin{figure}[!htbp]
\centering
\includegraphics[width=\linewidth]{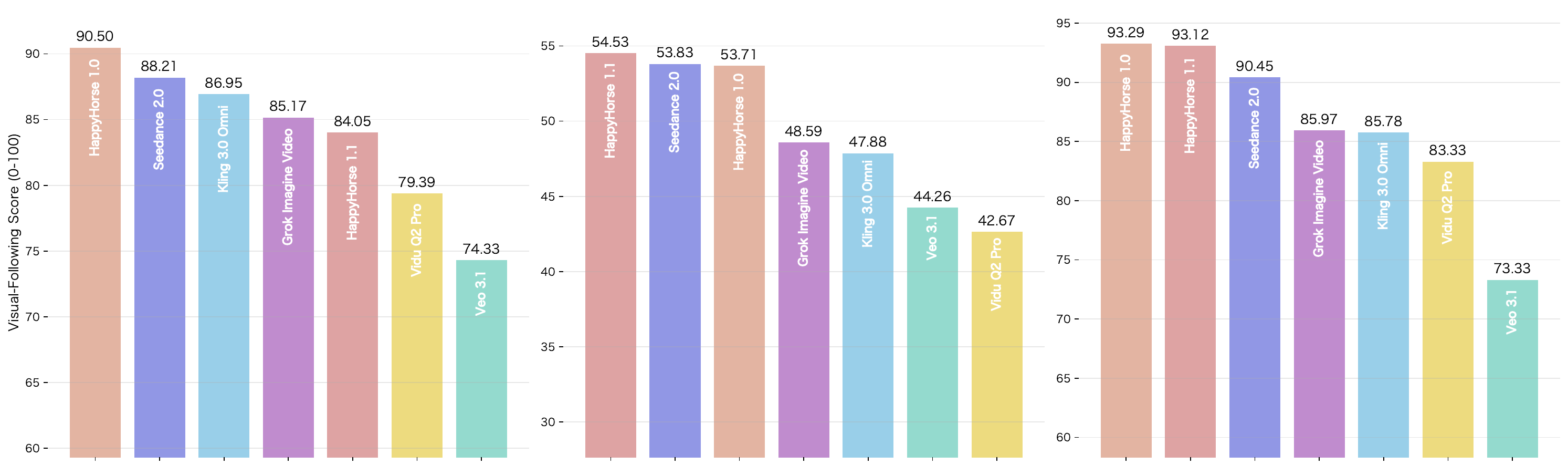}
\caption{R2V visual-following (reference-fidelity) ranking by reference type
(columns: scene / prop / character).}
\label{fig:reference}
\end{figure}

R2V prompts condition on one of three reference types: scene (232 prompts),
prop (209) and character (213). Figure~\ref{fig:reference} reports the
reference-fidelity (visual-following) ranking for each type.
The visual-following leader differs from the overall leader:
HappyHorse~1.0 tops scene- and character-fidelity, while HappyHorse~1.1 leads
on prop-fidelity, indicating that raw reference fidelity and overall
film-quality are distinct capabilities.
Across the three types, scene-space and character-appearance fidelity show
a similar ranking order, whereas prop-reference fidelity reveals a notably
different model ordering and a much tighter spread, confirming that
current models still struggle to faithfully preserve a referenced prop even
when scene and character references are respected.

\subsection{Cinematic Language Related L3 Landscape}
\label{sec:exp:bottleneck}

\begin{figure}[!htbp]
\centering
\includegraphics[width=\linewidth]{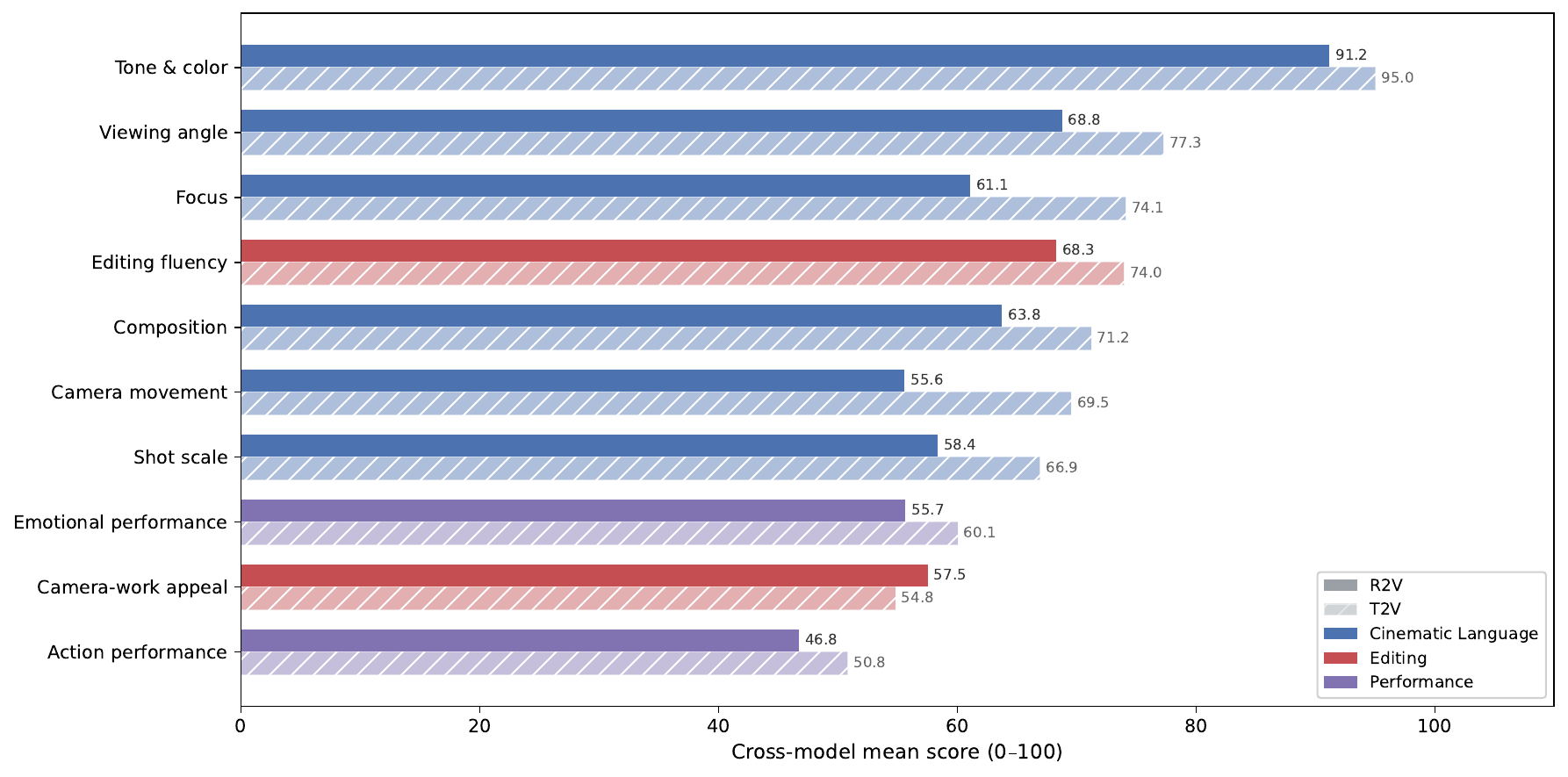}
\caption{Cinematic Language L3 sub-metric cross-model means (0--100), T2V
(hatched) vs.\ R2V (solid), sorted by descending score. Bars are colored by
L2 group: Cinematic Language (blue), editing (red), performance (purple).}
\label{fig:film_l3}
\end{figure}

Finally, we zoom in on the L3 sub-metrics most directly tied to film
production craft. Figure~\ref{fig:film_l3} collects 10 sub-metrics from
three L2 groups across the two tasks: Cinematic Language (camera movement, shot
scale, focus, viewing angle, tone \& color, composition), editing
(camera-work appeal, editing fluency), and performance (action performance,
emotional performance). Across both tasks, models score highest on tone \&
color (T2V 95.0, R2V 91.2) and lowest on action performance (50.8 / 46.8)
and camera-work appeal (54.8 / 57.5). The performance group is uniformly
the lowest-scoring L2 cluster in both tasks, while Cinematic Language dimensions
show the widest T2V--R2V gap (e.g., camera movement 69.5 vs.\ 55.6),
indicating that reference conditioning amplifies the spread on camera-work
execution.

\subsection{Summary of Findings}
\label{sec:exp:summary}

Taken together, the analyses above distill into five findings that hold across
the field. \textbf{(i)~No saturation; models differ most on Cinematic Language sub-metrics:} no model nears the ceiling on either task (tops of 88.93
/ 86.66), and camera movement, focus, shot scale and viewing angle show the
largest inter-model spread, far more so under R2V (camera-movement variance
$384.7$ vs.\ $135.7$), while temporal and audio dimensions show near-uniform
scores. \textbf{(ii)~Dynamic aesthetics are the lowest-scoring sub-metrics across the field:}
action performance, camera-work appeal, character-motion realism and emotional
performance are the lowest sub-metrics in both tasks, while static
image-quality sub-metrics (sharpness, lighting) score markedly higher. \textbf{(iii)~Multi-shot prompts are uniformly
harder and show wider model spread:} every model scores lower on multi-shot
prompts (average drop 7.9 points), with the drop larger for lower-ranked
models;
the drop concentrates on shot-craft dimensions (Cinematic Language, editing
fluency, scene space). \textbf{(iv)~Reference conditioning stresses
rather than reorders:} it leaves the top group intact but compresses the field,
and the visual-following leader differs from the overall leader, indicating
that raw reference fidelity and overall film-quality are distinct capabilities.
\textbf{(v)~No model wins on all L3 sub-metrics:} Seedance~2.0, the overall
leader, claims only 18/35 (T2V) or 20/38 (R2V) championships, concentrated on
Cinematic Language sub-metrics (Cinematic Language, editing appeal and performance),
while HappyHorse~1.1 dominates 11/35 (T2V) or 10/38 (R2V) sub-metrics on
character, audio and scene dimensions; notably the three R2V-specific
visual-following championships all go to the HappyHorse family, not to
Seedance; 4~(T2V) and 2~(R2V) models do not claim any championship. This ``championship mismatch''
reveals structural complementarity concealed by a single leaderboard number.

% !TeX root = neurips_2026.tex
% ===========================================================
% Section: Conclusion
% ===========================================================
\section{Conclusion}
\label{sec:conclusion}

We presented FilmBench, an evaluation benchmark for text-to-video (T2V) and
reference-to-video (R2V) generation grounded in the Cinematic Language
system used in professional film production. Built in collaboration with
directors and faculty from the Beijing Film Academy and a professional film
studio, FilmBench features prompts reverse-engineered from award-winning real
films, an academy-aligned three-level evaluation taxonomy (3 L1 axes, 12 L2
components, 35 L3 sub-metrics), and an expert-grade automatic evaluator
suite (FilmOps) whose model-level ranking reproduces expert rankings at
Spearman $\rho=0.95$ (T2V) / $0.96$ (R2V). We evaluated 9 T2V and 7 R2V
leading systems and released the benchmark to support research toward
film-grade generative systems.

\paragraph{Limitations.}
FilmBench is designed for evaluating professional film and cinematic content
creation capabilities. The evaluation dimensions and model rankings presented
in this work do not necessarily reflect model performance on non-cinematic
video generation tasks such as vertical short-form video creation,
casual user-generated content, or other non-professional video production
scenarios.

\paragraph{Broader Impact.}
We do not anticipate any negative societal impacts arising from this work.
FilmBench is an evaluation benchmark and does not itself generate or deploy
content; its intended use is to support research toward higher-quality video
generation systems.

% ----- Acknowledgments (hidden in anonymized submission; fill for camera-ready) -----
% \begin{ack}
% TODO: funding and competing-interests disclosure for the final paper.
% \end{ack}

\medskip
{\small
\bibliographystyle{plainnat}
\bibliography{references}
}

%%%%%%%%%%%%%%%%%%%%%%%%%%%%%%%%%%%%%%%%%%%%%%%%%%%%%%%%%%%%

\appendix

% !TeX root = neurips_2026.tex
% ===========================================================
% Appendix — reorganized A1→C2
% ===========================================================

% ===================== A1 =====================
\section{Qualitative Evaluation Examples}
\label{app:examples}

\begin{figure}[!htbp]
\centering
\includegraphics[width=\linewidth]{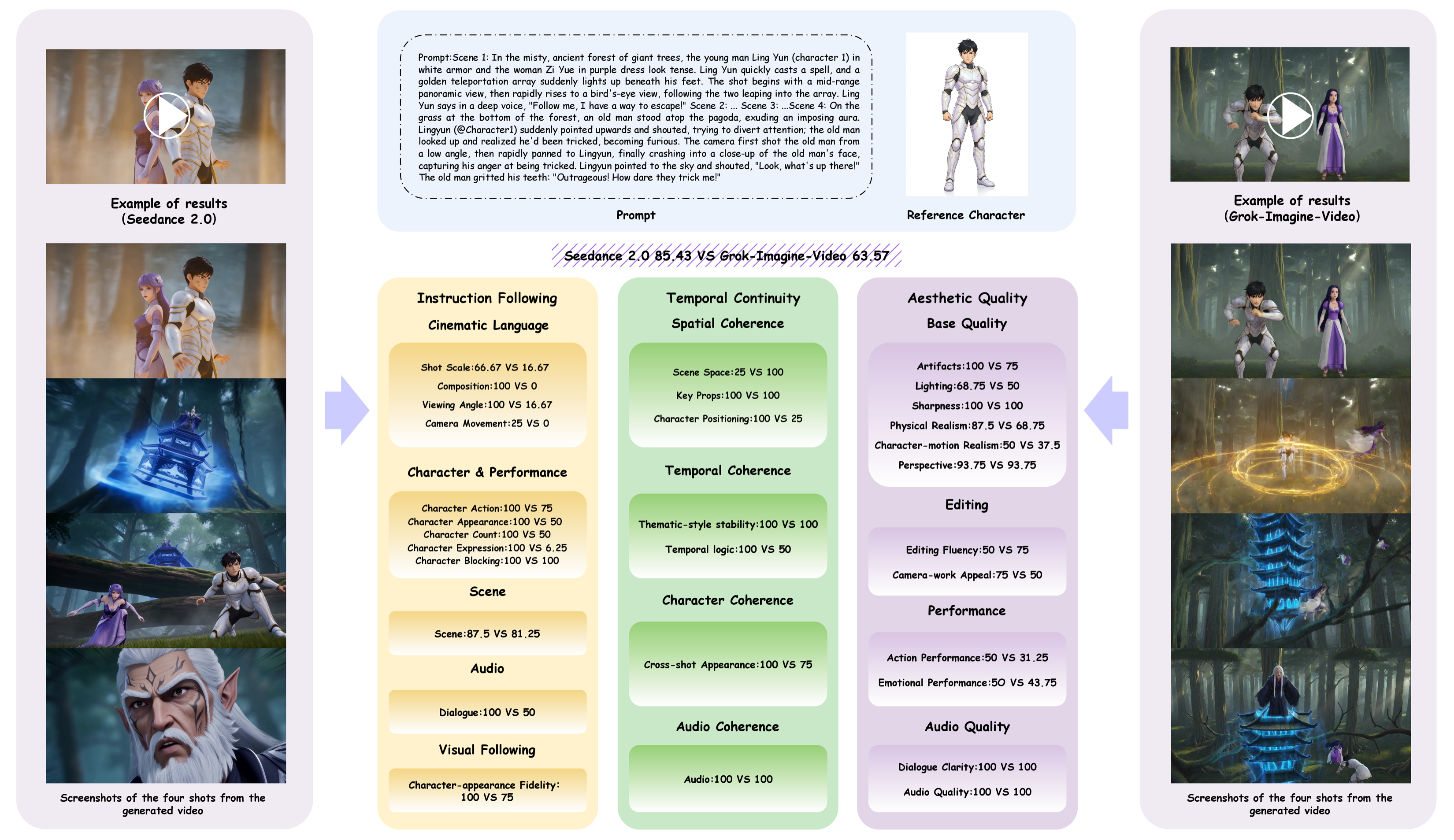}
\caption{R2V reference-character example in 3D animation: Seedance~2.0 (85.43) vs.\ Grok~Imagine~Video (63.57). Character-appearance fidelity is high, but the score gap is driven by Cinematic Language and character performance.}
\label{fig:eval_example_2}
\end{figure}

\begin{figure}[!htbp]
\centering
\includegraphics[width=\linewidth]{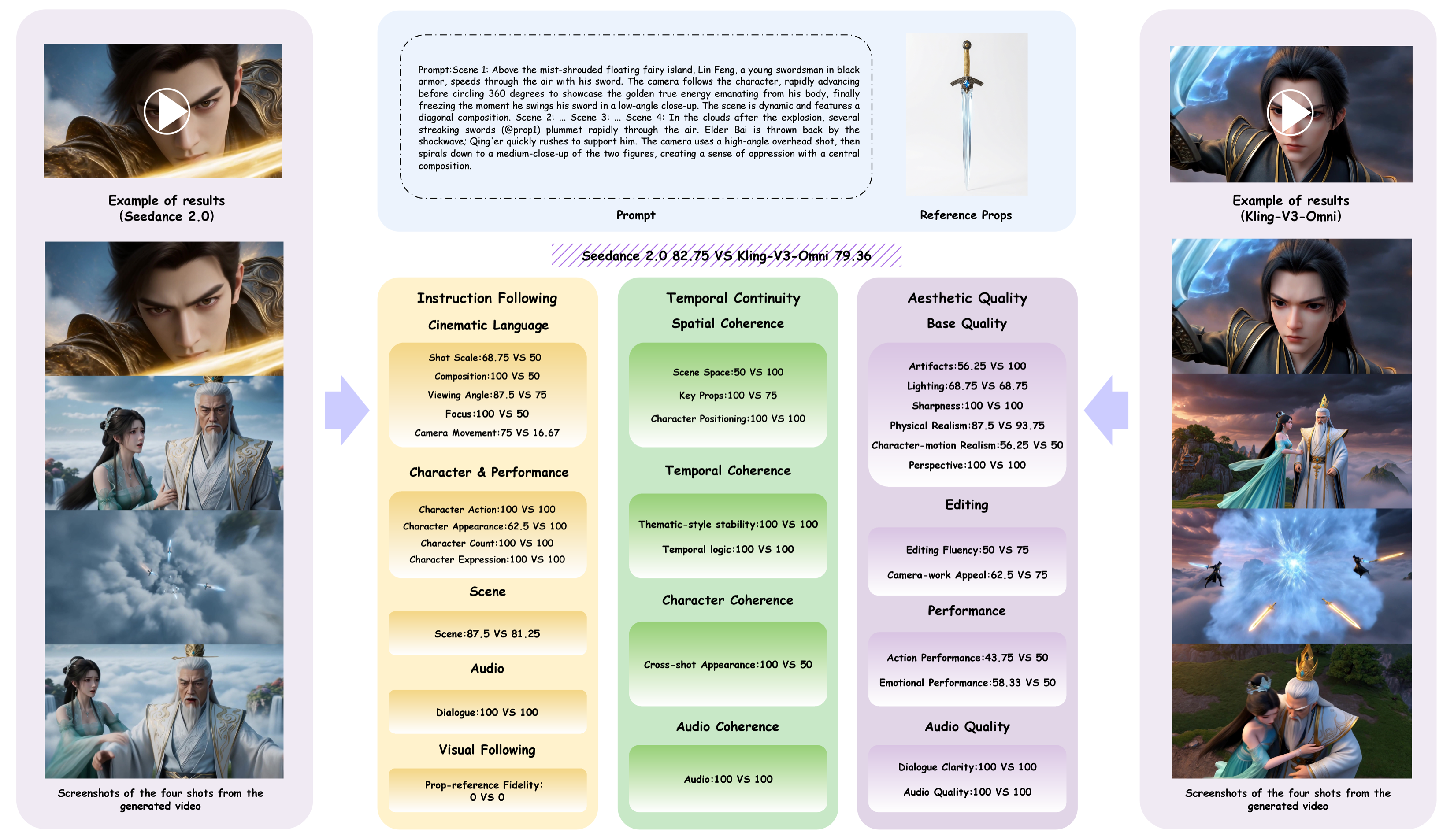}
\caption{R2V reference-prop example in 3D animation: Seedance~2.0 (82.75) vs.\ Kling~3.0~Omni (79.36). Both score 0 on prop-reference fidelity because the prop's color is altered by a golden energy aura intended for
the character; the overall gap is again driven by Cinematic Language.}
\label{fig:eval_example_3}
\end{figure}

Figures~\ref{fig:eval_example_2}--\ref{fig:eval_example_3} present two R2V
cases from 3D animated film production, illustrating how FilmBench's
fine-grained scoring reveals dimension-level trade-offs that an aggregate
score alone would conceal. In both cases, the primary score gaps concentrate
on Cinematic Language and character performance within Instruction Following,
rather than on temporal or aesthetic dimensions where models tend to cluster.

The reference-character case (Figure~\ref{fig:eval_example_2}) shows that
character-appearance fidelity scores are relatively high across models, as
the reference identity is generally well preserved. By contrast, the
reference-prop case (Figure~\ref{fig:eval_example_3}) exposes a field-wide
prop-reference bottleneck: both models score 0 on prop-reference fidelity
because the prompt describes a golden energy aura around the character's
body, yet the generated effect bleeds onto the reference prop and alters
its color, illustrating that current models struggle to isolate stylized
visual effects from unrelated reference objects.

% ===================== A2 =====================
\section{L3 Dimension Definitions}
\label{app:dims}

Tables~\ref{tab:l3defs} and~\ref{tab:l3defs2} list all 35 L3 sub-metrics
scored in the T2V task, grouped by their L1 axis and L2 component. The R2V task
adds three further L3
sub-metrics (\emph{scene-space}, \emph{character-appearance} and
\emph{prop-reference} fidelity) under a Visual-Following L2 component of
Instruction Following, each scored against the corresponding reference
type, giving 38 in total. Each sub-metric is rated on a 1--5 scale against
an anchor rubric.

\begin{table}[!htbp]
\caption{FilmBench L3 sub-metric definitions, part~1: the Instruction
Following (IF) axis, including the three R2V-only Visual-Following
sub-metrics. ``IF'' = Instruction Following, ``TC'' = Temporal
Continuity, ``AQ'' = Aesthetic Quality.}
\label{tab:l3defs}
\centering
\small
\begin{tabular}{p{1.0cm}p{2.4cm}p{2.6cm}p{6.6cm}}
\toprule
L1 & L2 component & L3 sub-metric & Definition (5-point rubric anchor) \\
\midrule
\multirow{18}{*}{IF}
 & \multirow{7}{2.4cm}{Cinematic Language}
   & Shot scale        & Adherence to requested shot scale (extreme close-up $\to$ extreme long shot). \\
 & & Camera movement   & Adherence to requested camera move (fixed, push, pull, pan, track, orbit, follow, zoom, handheld, strong move). \\
 & & Viewing angle     & Adherence to requested horizontal/vertical angle (eye-level, high, bird's-eye, extreme low/high, Dutch, over-shoulder, POV). \\
 & & Composition       & Adherence to requested composition rule (center, symmetry, rule-of-thirds, diagonal, leading-line, framing, depth, etc.). \\
 & & Focus             & Whether in/out-of-focus and focus changes match the prompt. \\
 & & Tone \& color     & Adherence to requested hue / temperature / saturation. \\
 & & Thematic style    & Adherence to requested style (realism, cyberpunk, horror; or a director's style). \\
\cmidrule(l){2-4}
 & \multirow{5}{2.4cm}{Character \& performance}
   & Character count      & Whether the number of characters matches the prompt. \\
 & & Character appearance & Whether facial features, hairstyle, accessories, age match the prompt. \\
 & & Character expression & Whether expression / emotion / gaze match, with no spurious expressions. \\
 & & Character action     & Whether actions match, with no spurious actions. \\
 & & Character blocking   & Whether blocking, entrance/exit and orientation match the prompt. \\
\cmidrule(l){2-4}
 & \multirow{2}{2.4cm}{Scene}
   & Scene           & Whether the scene matches the prompt. \\
 & & Fore/mid/background & Whether foreground/middle/background match the prompt. \\
\cmidrule(l){2-4}
 & \multirow{2}{2.4cm}{Audio}
   & Dialogue        & Dialogue completeness, emotional accuracy, lip-sync, no ID jumps. \\
 & & Sound effects   & Whether sound effects / BGM match the prompt. \\
\cmidrule(l){2-4}
 & \multirow{3}{2.4cm}{\textit{Visual Following (R2V)}}
   & \textit{Scene-space fidelity}       & \textit{Whether the generated space/scene stays consistent with the reference scene image (scene reference).} \\
 & & \textit{Character-appearance fidelity} & \textit{Whether character appearance stays consistent with the reference character image (character reference).} \\
 & & \textit{Prop-reference fidelity}     & \textit{Whether key props stay consistent with the reference prop image (prop reference).} \\
\bottomrule
\end{tabular}
\end{table}

\begin{table}[!htbp]
\caption{FilmBench L3 sub-metric definitions, part~2: the Temporal Continuity
(TC) and Aesthetic Quality (AQ) axes.}
\label{tab:l3defs2}
\centering
\small
\begin{tabular}{p{1.0cm}p{2.4cm}p{2.6cm}p{6.6cm}}
\toprule
L1 & L2 component & L3 sub-metric & Definition (5-point rubric anchor) \\
\midrule
\multirow{7}{*}{TC}
 & \multirow{3}{2.4cm}{Spatial coherence}
   & Scene space       & Cross-shot spatial/layout consistency; no unexplained jumps or appearing/vanishing elements. \\
 & & Key props         & Key props keep count, form, material and color; no flicker or vanishing. \\
 & & Character positioning & Character positioning/orientation stays temporally consistent. \\
\cmidrule(l){2-4}
 & \multirow{2}{2.4cm}{Temporal coherence}
   & Temporal logic    & Events unfold with coherent causal/temporal logic, no fragmentation. \\
 & & Thematic-style stability & Style / tone / grading stay unified over time, no drift. \\
\cmidrule(l){2-4}
 & Character coherence & Cross-shot appearance & Same character keeps face and overall appearance throughout, no ID drift. \\
\cmidrule(l){2-4}
 & Audio coherence & Audio & Timbre / volume / audio-style stay consistent over time. \\
\midrule
\multirow{12}{*}{AQ}
 & \multirow{6}{2.4cm}{Base quality}
   & Sharpness              & Image clarity and detail resolution. \\
 & & Physical realism       & Props/scene obey real-world physics (excluding character motion). \\
 & & Character-motion realism & Character movement is natural and physically plausible; no distortion / clipping. \\
 & & Artifacts              & Free of AI artifacts (flicker, noise, extra limbs, garbled text, seams). \\
 & & Perspective            & Correct scaling, geometry and spatial consistency; no perspective conflict. \\
 & & Lighting               & Correct exposure, consistent light/shadow, atmosphere and dimensionality. \\
\cmidrule(l){2-4}
 & \multirow{2}{2.4cm}{Editing}
   & Editing fluency        & Smooth multi-shot editing: subject continuity, shot-scale/axis consistency, eyeline match, pacing. \\
 & & Camera-work appeal     & Whether camera work is compelling, purposeful and serves the narrative. \\
\cmidrule(l){2-4}
 & \multirow{2}{2.4cm}{Performance}
   & Emotional performance  & Expressiveness and believability of emotional acting. \\
 & & Action performance     & Quality and believability of physical/action performance. \\
\cmidrule(l){2-4}
 & \multirow{2}{2.4cm}{Audio quality}
   & Audio quality          & Overall audio production quality. \\
 & & Dialogue clarity       & Intelligibility and clarity of dialogue. \\
\bottomrule
\end{tabular}
\end{table}

% ===================== B1 =====================
\section{Per-Task Variance Breakdown}
\label{app:variance}

\begin{figure}[!htbp]
\centering
\includegraphics[width=\linewidth]{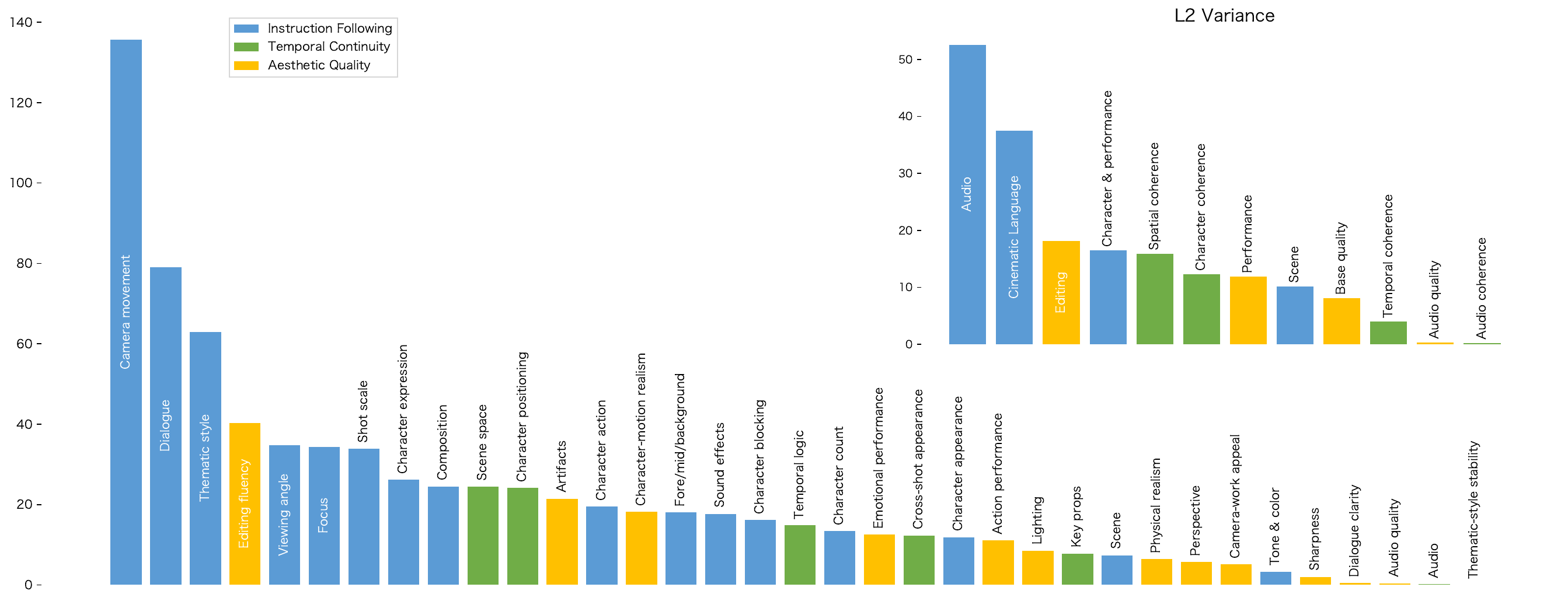}
\caption{T2V cross-model variance per L3 sub-metric, with L2 inset (upper right).}
\label{fig:var_t2v}
\end{figure}

\begin{figure}[!htbp]
\centering
\includegraphics[width=\linewidth]{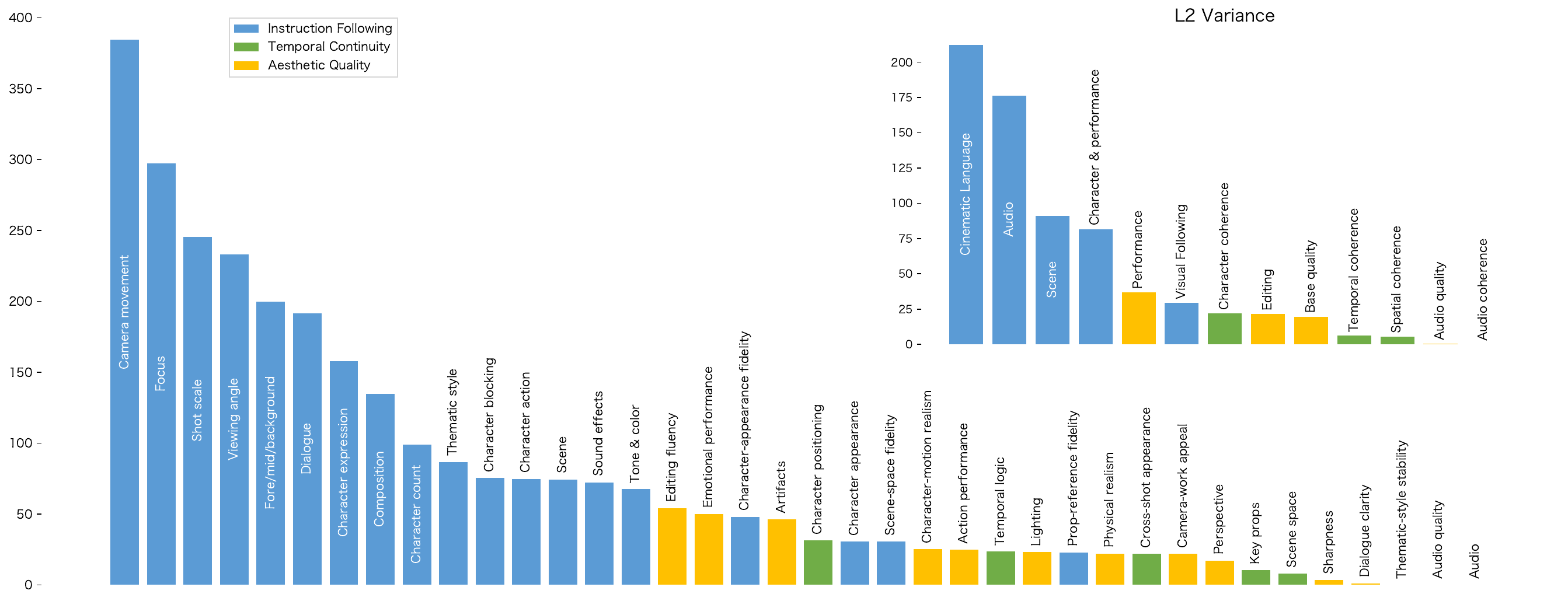}
\caption{R2V cross-model variance per L3 sub-metric, with L2 inset (upper right).}
\label{fig:var_r2v}
\end{figure}

Figures~\ref{fig:var_t2v} and~\ref{fig:var_r2v} give the per-task
cross-model variance for each L3 sub-metric.  The merged view combining
both tasks is shown in the main text (Figure~\ref{fig:var_combined}).

% ===================== B2 =====================
\section{Degradation Analysis}
\label{app:degradation}

Tables~\ref{tab:shot_dims} and~\ref{tab:action_dims} detail the performance
drops observed in single- to multi-shot and dialogue-to-action transitions,
supporting the analysis in Section~\ref{sec:exp:content}.

\begin{table}[!htbp]
\caption{T2V multi-shot performance degradation by taxonomy level: the worst-hit L1 axis, the
three sharpest L2 components and the five sharpest L3 sub-metrics. Values are mean
scores (0--100) on single- vs multi-shot prompts, macro-averaged over the nine
models; $\Delta$ is the multi-shot drop.}
\label{tab:shot_dims}
\centering
\small
\begin{tabular}{llccc}
\toprule
Level & Dimension & Single & Multi & $\Delta$ \\
\midrule
L1 & Instruction Following & 91.9 & 80.4 & $-11.5$ \\
   & Temporal Continuity   & 97.8 & 91.0 & $-6.8$ \\
   & Aesthetic Quality     & 77.7 & 72.4 & $-5.3$ \\
\midrule
L2 & Cinematic Language        & 86.6 & 69.4 & $-17.2$ \\
   & Editing               & 75.0 & 61.5 & $-13.5$ \\
   & Spatial coherence     & 96.0 & 84.5 & $-11.5$ \\
\midrule
L3 & Editing fluency       & 97.3 & 67.5 & $-29.8$ \\
   & Focus                 & 91.7 & 68.7 & $-23.0$ \\
   & Camera movement       & 87.0 & 65.1 & $-21.9$ \\
   & Viewing angle         & 88.6 & 74.3 & $-14.3$ \\
   & Scene space             & 93.4 & 80.1 & $-13.3$ \\
\bottomrule
\end{tabular}
\end{table}

\begin{table}[!htbp]
\caption{T2V action performance degradation by taxonomy level: the worst-hit L1 axis and the
sharpest L3 sub-metrics, split into a camera-work cluster (Instruction
Following) and a physical-realism cluster (Aesthetic Quality). Values are mean
scores (0--100) on dialogue vs.\ action prompts, macro-averaged over the nine
models; $\Delta$ is the action drop.}
\label{tab:action_dims}
\centering
\small
\begin{tabular}{llccc}
\toprule
Level & Dimension & Dialogue & Action & $\Delta$ \\
\midrule
L1 & Instruction Following & 85.7 & 73.5 & $-12.2$ \\
   & Aesthetic Quality     & 75.9 & 65.5 & $-10.4$ \\
\midrule
L3 & Camera movement       & 76.5 & 45.4 & $-31.1$ \\
   & Focus                 & 81.5 & 52.9 & $-28.6$ \\
   & Shot scale            & 72.9 & 46.3 & $-26.6$ \\
   & Physical plausibility & 86.3 & 69.3 & $-16.9$ \\
   & Character-motion realism & 61.8 & 45.9 & $-15.8$ \\
   & Perspective           & 94.5 & 79.2 & $-15.3$ \\
   & Artifacts             & 67.6 & 55.7 & $-11.9$ \\
\bottomrule
\end{tabular}
\end{table}

% ===================== B3 =====================
\section{Per-axis Scores}
\label{app:pillar}

This appendix reports per-model breakdowns for T2V and R2V under the
sample-level \emph{stand} aggregation used throughout the paper (machine
scores over the full prompt set; T2V $N=515$, R2V $N=654$). Model
abbreviations: Seed (Seedance~2.0), HH1.1/HH1.0 (HappyHorse~1.1/1.0), KV3O
(Kling~3.0~Omni), KV3 (Kling~3.0), Grok (Grok~Imagine~Video), Veo (Veo~3.1),
ViduQ3/ViduQ2 (Vidu~Q3-Pro/Q2-Pro), Hailuo (Hailuo~2.3).

Tables~\ref{tab:t2v-pillar}--\ref{tab:r2v-visual} give the per-axis breakdown
for both tasks, complementing the overall rankings in the main text.

\begin{table}[!htbp]
\caption{Per-axis scores (0--100) by model, T2V ($N=515$). Overall is the
equal-weighted mean of the three axes.}
\label{tab:t2v-pillar}
\centering
\small
\begin{tabular}{lcccc}
\toprule
Model & \textbf{Overall} & Instruction & Temporal & Aesthetic \\
\midrule
Seedance 2.0        & \textbf{88.9} & 91.7 & 96.1 & 79.0 \\
HappyHorse 1.1      & \textbf{87.4} & 90.7 & 94.8 & 76.8 \\
HappyHorse 1.0      & \textbf{87.0} & 89.5 & 94.5 & 76.9 \\
Kling 3.0 Omni       & \textbf{86.1} & 88.3 & 94.7 & 75.1 \\
Kling 3.0           & \textbf{85.8} & 87.5 & 94.6 & 75.2 \\
Vidu Q3 Pro         & \textbf{80.9} & 82.0 & 90.1 & 70.6 \\
Grok Imagine Video  & \textbf{80.9} & 82.0 & 89.0 & 71.6 \\
Veo 3.1             & \textbf{80.8} & 78.5 & 90.4 & 73.6 \\
Hailuo 2.3          & \textbf{68.9} & 55.9 & 88.1 & 62.7 \\
\bottomrule
\end{tabular}
\end{table}

\begin{table}[!htbp]
\caption{Per-axis scores (0--100) by model, R2V ($N=654$).}
\label{tab:r2v-pillar}
\centering
\small
\begin{tabular}{lcccc}
\toprule
Model & \textbf{Overall} & Instruction & Temporal & Aesthetic \\
\midrule
Seedance 2.0        & \textbf{86.7} & 87.5 & 94.1 & 78.4 \\
HappyHorse 1.1      & \textbf{85.5} & 86.7 & 93.3 & 76.5 \\
HappyHorse 1.0      & \textbf{84.9} & 85.0 & 93.4 & 76.2 \\
Kling 3.0 Omni       & \textbf{80.4} & 78.8 & 90.5 & 71.9 \\
Veo 3.1             & \textbf{78.2} & 74.7 & 89.2 & 70.8 \\
Grok Imagine Video  & \textbf{76.9} & 71.2 & 90.3 & 69.3 \\
Vidu Q2 Pro         & \textbf{71.4} & 57.4 & 88.1 & 68.8 \\
\bottomrule
\end{tabular}
\end{table}

\begin{table}[!htbp]
\caption{R2V Visual-Following fidelity (0--100) by reference type. Each R2V
prompt is scored only on the sub-metric matching its reference type (scene
$N=232$, character $N=213$, prop $N=209$), so columns report scores on
disjoint prompt subsets.}
\label{tab:r2v-visual}
\centering
\small
\begin{tabular}{lccc}
\toprule
Model & Scene-space & Character-appearance & Prop-reference \\
\midrule
Seedance 2.0        & 88.2 & 90.5 & 53.8 \\
HappyHorse 1.1      & 84.0 & 93.1 & 54.5 \\
HappyHorse 1.0      & 90.5 & 93.3 & 53.7 \\
Kling 3.0 Omni       & 86.9 & 85.8 & 47.9 \\
Veo 3.1             & 74.3 & 73.3 & 44.3 \\
Grok Imagine Video  & 85.2 & 86.0 & 48.6 \\
Vidu Q2 Pro         & 79.4 & 83.3 & 42.7 \\
\bottomrule
\end{tabular}
\end{table}

Table~\ref{tab:t2v-market} summarizes T2V robustness across markets: the top
ranks are stable across Chinese and international source titles
(Seedance~2.0 is \#1 in both).

\begin{table}[!htbp]
\caption{T2V overall scores (0--100) on the market-labeled prompt subset.
Columns restrict to prompts drawn from Chinese ($1{,}431$ records) vs.\
international ($1{,}764$ records) source titles; prompts without a market
label are excluded.}
\label{tab:t2v-market}
\centering
\small
\begin{tabular}{lcc}
\toprule
Model & CN & INTL \\
\midrule
Seedance 2.0        & 90.3 & 90.4 \\
HappyHorse 1.1      & 89.1 & 89.5 \\
HappyHorse 1.0      & 89.1 & 89.1 \\
Kling 3.0 Omni       & 88.1 & 88.7 \\
Kling 3.0           & 87.7 & 88.7 \\
Vidu Q3 Pro         & 84.2 & 84.2 \\
Grok Imagine Video  & 84.4 & 85.2 \\
Veo 3.1             & 84.5 & 84.6 \\
Hailuo 2.3          & 73.6 & 74.3 \\
\bottomrule
\end{tabular}
\end{table}

% ===================== B4 =====================
\section{Full L3 / L2 Heatmaps}
\label{app:heatmaps}

\begin{figure}[!htbp]
\centering
\includegraphics[width=\linewidth,height=0.92\textheight,keepaspectratio]{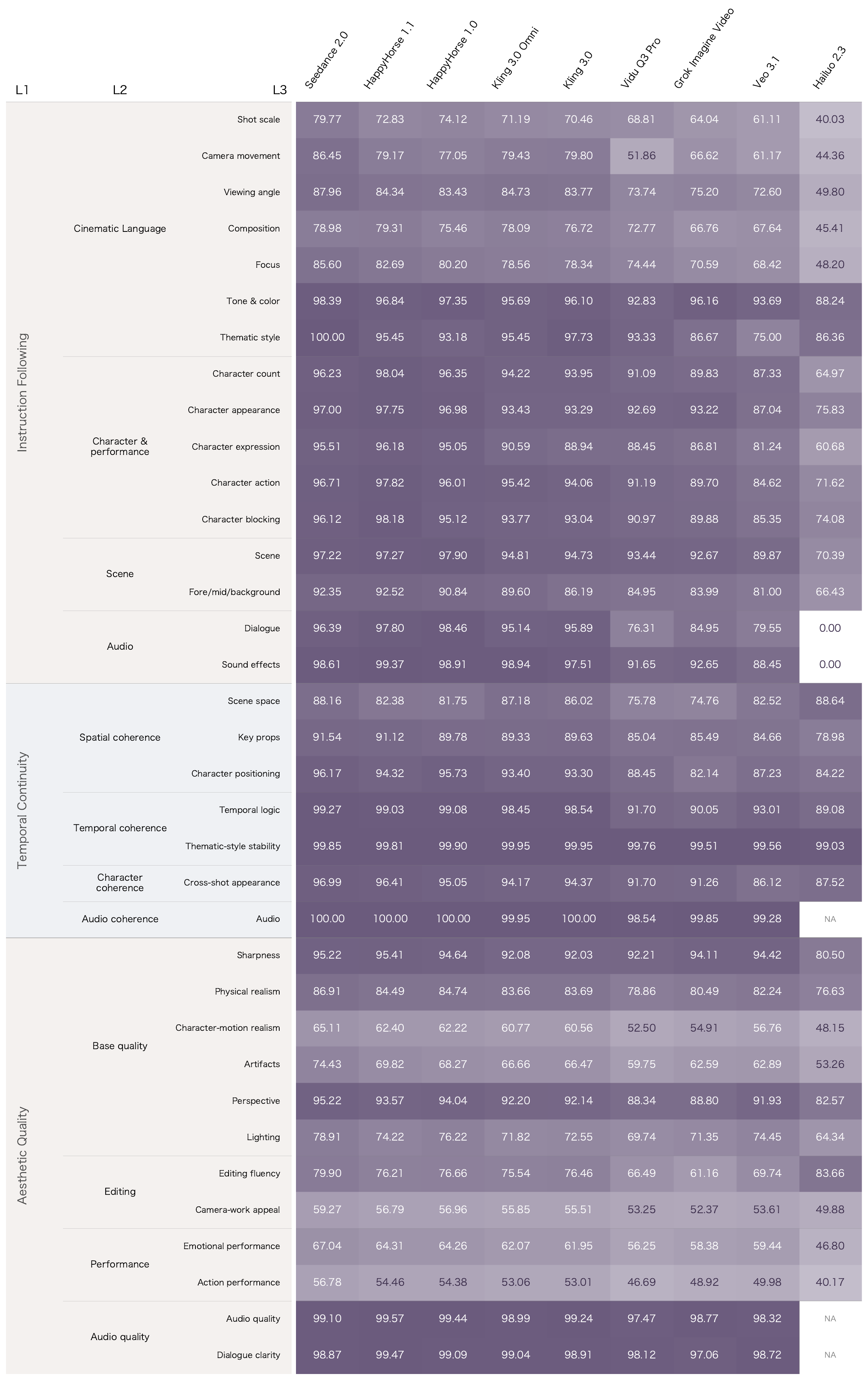}
\caption{T2V per-model L3 heatmap (35 sub-metrics $\times$ 9 models).}
\label{fig:heat_l3_t2v}
\end{figure}

\begin{figure}[!htbp]
\centering
\includegraphics[width=\linewidth,height=0.92\textheight,keepaspectratio]{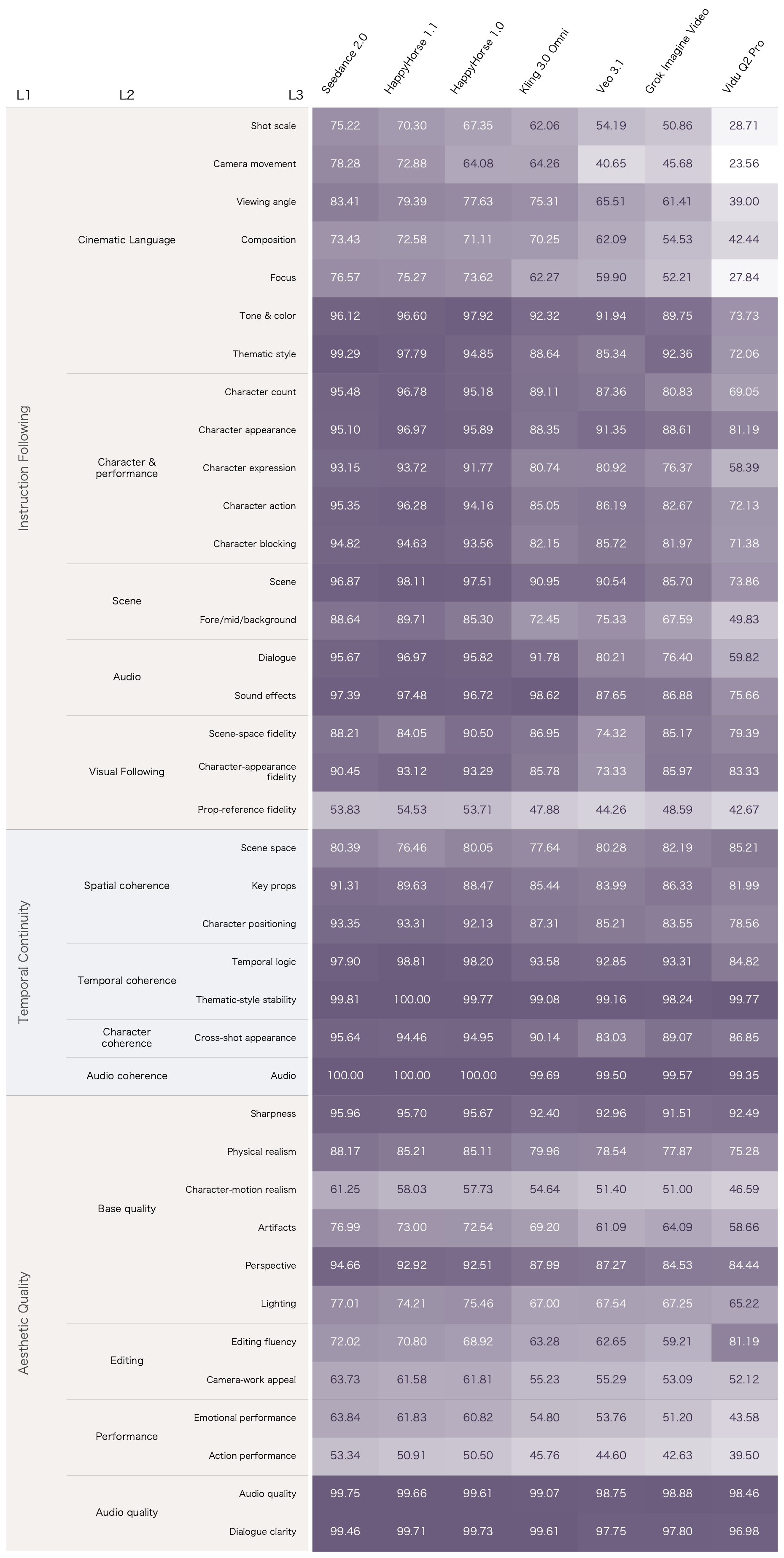}
\caption{R2V per-model L3 heatmap (38 sub-metrics $\times$ 7 models; adds the
three Visual Following fidelity rows).}
\label{fig:heat_l3_r2v}
\end{figure}

\begin{figure}[!htbp]
\centering
\includegraphics[width=0.9\linewidth]{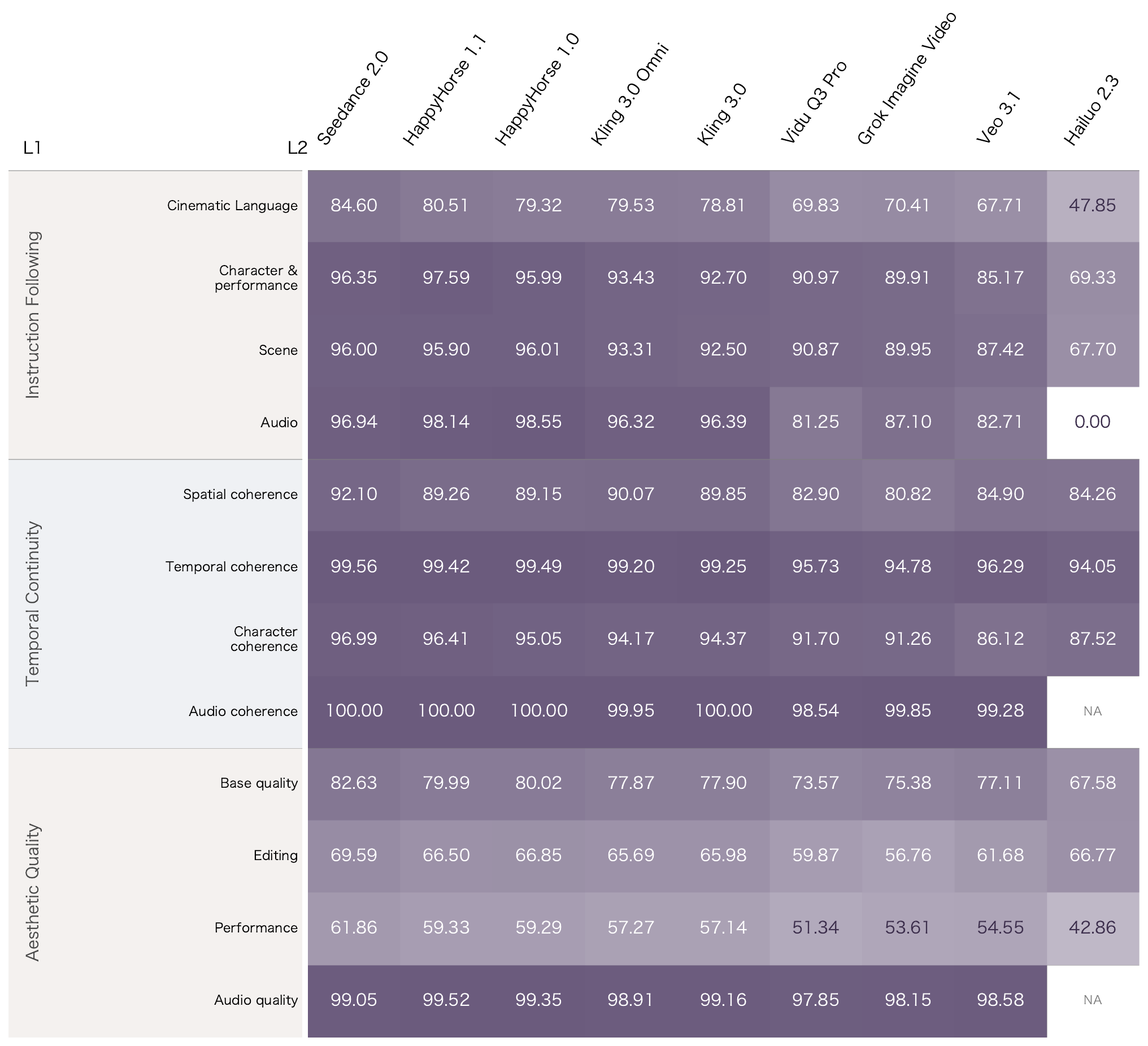}
\caption{T2V per-model L2 heatmap (12 components $\times$ 9 models).}
\label{fig:heat_l2_t2v}
\end{figure}

\begin{figure}[!htbp]
\centering
\includegraphics[width=0.9\linewidth]{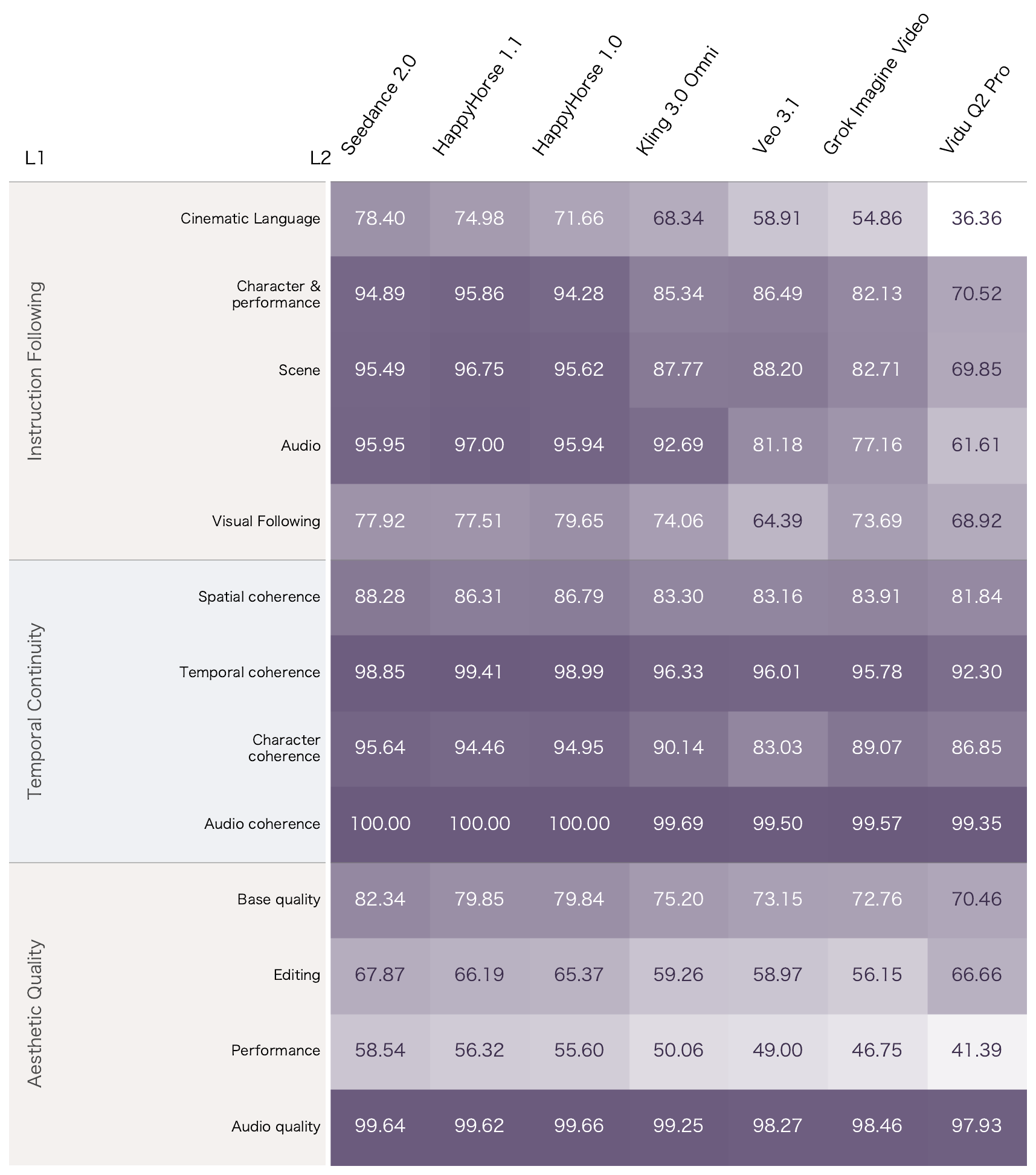}
\caption{R2V per-model L2 heatmap (13 components $\times$ 7 models; adds the
Visual Following component).}
\label{fig:heat_l2_r2v}
\end{figure}

For completeness, Figures~\ref{fig:heat_l3_t2v}--\ref{fig:heat_l2_r2v} give
the complete per-model score matrices at both granularities. Rows are grouped
by L1 axis (and L2 component for the L3 maps); columns are ordered by overall
rank. Darker cells are higher scores. These maps make the low-variance
rows (temporal/audio) and the discriminative rows (camera/framing, dynamic
aesthetics, and, for R2V, prop-reference fidelity) visible at a glance, and
corroborate the aggregated rankings in the main text.

% ===================== C1 =====================
\section{FilmOps Taxonomy Details}
\label{app:filmops:taxonomy}

FilmOps grounds the Cinematic Language dimensions by mapping each frame or
shot into structured cinematic labels. Its taxonomy spans six core
dimensions and 55 fixed sub-categories (character layout is an open-ended
natural-language field and is not counted), with all category definitions
aligned to classical production references and validated by practitioners.
Table~\ref{tab:filmops_tax} summarizes the dimensions, and the category
inventories are listed below.

\begin{table}[!htbp]
\caption{FilmOps taxonomy: six operators, their granularity, number of label
classes and label form. Total: 55 fixed sub-categories (character layout is an
open-ended natural-language field, not counted).}
\label{tab:filmops_tax}
\centering
\small
\begin{tabular}{llcl}
\toprule
Dimension & Granularity & \# Classes & Label form \\
\midrule
Shot scale       & Frame & 8  & single-label \\
Composition      & Frame & 12 & multi-label \\
Viewing angle    & Frame & 7  & multi-label \\
Tone \& color     & Frame & 18 & multi-label (3 sub-axes) \\
Character layout & Frame & -- & natural language \\
Camera movement  & Shot  & 10 & multi-label \\
\bottomrule
\end{tabular}
\end{table}

\paragraph{Category inventories.}
\textbf{Shot scale} (8): extreme close-up, close-up, close shot, medium shot,
medium full shot, full shot, long shot, extreme long shot.
\textbf{Composition} (12): center, rule of thirds, horizontal, vertical,
symmetric, framing, scattered, leading lines, diagonal, oblique, triangular,
depth of field.
\textbf{Viewing angle} (7): eye level, low angle, high angle, bird's eye,
extreme low angle, extreme high angle, Dutch angle.
\textbf{Tone \& color} (18, three sub-axes): hue (red, orange, yellow, green,
cyan, blue, purple, magenta, pink, brown, monochrome, white); temperature
(cool, warm, mixed); and saturation (high, medium, low).
\textbf{Character layout}: an open-ended natural-language description of each
main character's on-screen position and orientation, expressed in image-frame
coordinates.
\textbf{Camera movement} (10): push in, pull out, pan, tracking, static, arc,
follow, roll, zoom, dynamic (strong movement).

% ===================== C2 =====================
\section{FilmOps Operator Evaluation}
\label{app:operators}

Table~\ref{tab:operators} reports, for each FilmOps operator, its backbone and
its accuracy against four strong zero-shot general MLLM baselines (GPT-4.1,
Qwen3-VL-235B, Gemini~3.5~Flash and Gemini~3.1~Pro). Classification operators
are scored by macro-F1 and the natural-language character-layout operator by
precision. The trained operators outperform every baseline on all six
dimensions, most sharply on the temporally/professionally demanding ones
(camera movement, composition, tone \& color), corroborating the design
discussion in the main text (Section~\ref{sec:eval:ops}).

\begin{table}[!htbp]
\caption{FilmOps operators vs.\ four zero-shot general MLLM baselines.
Classification operators are scored by macro-F1; the natural-language
character-layout operator by precision (``--'' denotes an unsupported
setting). "F" and "P" stand for "Flash" and "Pro".}
\label{tab:operators}
\centering
\small
\setlength{\tabcolsep}{4pt}
\resizebox{\textwidth}{!}{%
\begin{tabular}{llccccc}
\toprule
Operator & Backbone & GPT-4.1 & Qwen3-VL & Gemini~3.5~F & Gemini~3.1~P & FilmOps \\
\midrule
Shot scale        & DINO (ViT-B)    & 0.483 & 0.615 & 0.733 & 0.725 & \textbf{0.906} \\
Composition       & DINO (ViT-B)    & 0.330 & 0.289 & 0.370 & 0.500 & \textbf{0.859} \\
Viewing angle     & BEiT-Base       & 0.497 & 0.452 & 0.534 & 0.560 & \textbf{0.810} \\
Tone \& color     & ResNet-18       & 0.522 & 0.503 & 0.512 & 0.565 & \textbf{0.828} \\
Character layout  & InternVL3-14B    & 0.569 & 0.556 & 0.698 & 0.692 & \textbf{0.881} \\
Camera movement   & InternVL3-14B    & --    & 0.277 & 0.318 & 0.324 & \textbf{0.795} \\
\bottomrule
\end{tabular}%
}
\end{table}

\end{document}